\documentclass{article}

\newcommand{\squeezeup}{\vspace{-2.5mm}}
\usepackage{amsmath,amsfonts,cancel,multirow,subfigure,pifont,tikz}
\usetikzlibrary{arrows,shapes}
\usepackage{enumitem}
\usepackage{longtable}

\newcommand{\figref}[1]{Fig. \ref{#1}}
\newcommand{\secref}[1]{\S \ref{#1}}
\renewcommand{\eqref}[1]{Eq. (\ref{#1})}
\newcommand\independent{\protect\mathpalette{\protect\independenT}{\perp}} 
\def\independenT#1#2{\mathrel{\rlap{$#1#2$}\mkern2mu{#1#2}}}
\newcommand\ci{\independent}
\newcommand{\av}[1]{\langle #1 \rangle}

\definecolor{myred}{cmyk}{0,0.9,0.9,0.1}
\newcommand{\myred}{\color{myred}}

\tikzstyle{disc}=[rectangle,draw=blue!50,thick,line width=1pt,minimum size=6mm]  
\tikzstyle{obs}=[fill=blue!20,thick]  
\tikzstyle{hcont}=[ellipse,draw=black!50,thick,minimum size=6mm,>=stealth]  
\tikzstyle{ocont}=[ellipse,draw=black!100,thick,minimum size=6mm,>=stealth]  
\tikzstyle{dgraph}=[->, line width=1.5pt]

\newcommand{\ie}{\emph{i.e.}}
\newcommand{\eg}{\emph{e.g.}}

\usepackage{multicol}
\usepackage{hyperref}
\usepackage{microtype}
\usepackage{graphicx}
\usepackage{booktabs} 

\usepackage[accepted]{icml2018}

\icmltitlerunning{Path-Specific Counterfactual Fairness}

\usepackage[multiple]{footmisc}
\begin{document}

\twocolumn[
\icmltitle{Path-Specific Counterfactual Fairness}

\begin{icmlauthorlist}
\icmlauthor{Silvia Chiappa}{goo}
\icmlauthor{Thomas P. S. Gillam}{goo}
\end{icmlauthorlist}

\icmlaffiliation{goo}{DeepMind, London, UK}

\icmlcorrespondingauthor{Silvia Chiappa}{csilvia@google.com}

\icmlkeywords{Machine Learning, ICML}

\vskip 0.3in
]

\printAffiliationsAndNotice{}  

\begin{abstract}
We consider the problem of learning fair decision systems in complex scenarios in which a sensitive attribute might affect the decision along both fair and unfair pathways.
We introduce a causal approach to disregard effects along unfair pathways that simplifies and generalizes previous literature.
Our method corrects observations adversely affected by the sensitive attribute, and uses these to form a decision.
This avoids disregarding fair information, and does not require an often intractable computation of the path-specific effect. 
We leverage recent developments in deep learning and approximate inference to achieve a solution that is widely applicable to complex, non-linear scenarios.
\end{abstract}

\section{Introduction}
\label{intro}
Machine learning is increasingly being used to take decisions that can severely affect people's lives, \eg~in policing, education, hiring \cite{hoffman15discretion}, lending, and criminal risk assessment \cite{dieterich16compas}. 
However, most often the training data contains bias that exists in our society. This bias can be absorbed or even amplified by the systems,
leading to decisions that are unfair with respect to \emph{sensitive attributes} (\eg~race and gender).

In response to calls from governments and institutions, the research community has recently started to
address the issue of fairness through a variety of perspectives and frameworks. 
The simplest solution to this challenge is to down-weight or discard the sensitive attribute \cite{zeng16interpretable}. 
This can adversely impact model accuracy, and most often does not result in a fair procedure as the sensitive attribute might be correlated with the other attributes. 
A more sophisticated approach is to pre-process the data or extract representations that do not contain information about the sensitive attribute \cite{zemel13learning,feldman15certifying,edwards16censoring,louizos16fair,calmon17optimized}. 
Both types of approach  assume that the influence of the sensitive attribute on the decision is entirely unfair. 

In an attempt to formalize different notions of fairness, the community has introduced several statistical criteria for establishing whether a decision system is fair \cite{dwork12fairness,feldman15certifying,chouldechova17fair,corbett-davies17algorithmic},
and algorithms have been developed to attain a given fairness criterion by imposing constraints into the optimization, or by using an adversary \cite{zafar17fairness,zhang18mitigating}.
However, it is often unclear which criterion is most appropriate for a given decision problem. Even more problematic is the fact that criteria that intuitively seem to correspond to similar types of fairness 
cannot always be concurrently satisfied on a dataset \cite{kleinberg16inherent,berk17fairness,chouldechova17fair}. 
Finally, approaches based on statistical relations among observations are in danger of not discerning correlation from causation, and are unable to distinguish the different ways in which the sensitive 
attribute might influence the decision.

It was recently argued \cite{qureshi16causal,bonchi17exposing,kilbertus17avoiding,kusner17counterfactual,nabi18fair,russell17when,zhang17causal,zhang17anti,zhang18fairness} 
that using a causal framework \cite{pearl00causality,dawid07fundamentals,pearl16causal,peters16elements} would lead to a more intuitive, powerful, and less error-prone way of reasoning about fairness.
The suggestion is to view unfairness as the presence of an unfair \emph{causal effect} of the sensitive attribute on the decision, as already done in \citet{pearl00causality} and \citet{pearl16causal}, \eg, to analyse the case of Berkeley's alleged sex bias in graduate admission.

\citet{kusner17counterfactual} recently introduced a causal definition of fairness, called \emph{counterfactual fairness}, which states that a decision is fair toward an individual 
if it coincides with the one that would have been taken in a counterfactual world in which the sensitive attribute were different, and suggested a general algorithm to achieve this notion.
This definition considers the entire effect of the sensitive attribute on the decision as problematic. However, in many practical scenarios this is not the case.
For example, in the Berkeley alleged sex bias case, female applicants were rejected more often than male applicants as they were more often applying to departments with lower admission rates.
Such an effect of gender through department choice is not unfair.

To deal with such scenarios, we propose a novel fairness definition called \emph{path-specific counterfactual fairness}, which states that
a decision is fair toward an individual if it coincides with the one that would have been taken in a counterfactual world in which the sensitive attribute \emph{along the unfair pathways} were different. 

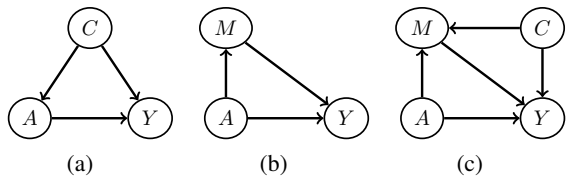
\begin{figure}[t]
\begin{center}
\subfigure[]{
\scalebox{0.8}{
\begin{tikzpicture}[dgraph]
\node[ocont] (C) at (1,1.5) {$C$};
\node[ocont] (A) at (0,0) {$A$};
\node[ocont] (Y) at (2,0) {$Y$};
\draw[line width=1.15pt](C)--(A);\draw[line width=1.15pt](C)--(Y);\draw[line width=1.15pt](A)--(Y);
\end{tikzpicture}}}
\hskip0.1cm
\subfigure[]{
\scalebox{0.8}{
\begin{tikzpicture}[dgraph]
\node[ocont] (M) at (0,1.5) {$M$};
\node[ocont] (X) at (0,0) {$A$};
\node[ocont] (Y) at (2,0) {$Y$};
\draw[line width=1.15pt](X)--(M);\draw[line width=1.15pt](X)--(Y);\draw[line width=1.15pt](M)--(Y);
\end{tikzpicture}}}
\hskip0.1cm
\subfigure[]{
\scalebox{0.8}{
\begin{tikzpicture}[dgraph]
\node[ocont] (C) at (2,1.5) {$C$};
\node[ocont] (M) at (0,1.5) {$M$};
\node[ocont] (X) at (0,0) {$A$};
\node[ocont] (Y) at (2,0) {$Y$};
\draw[line width=1.15pt](C)--(M);\draw[line width=1.15pt](C)--(Y);
\draw[line width=1.15pt](X)--(M);\draw[line width=1.15pt](X)--(Y);\draw[line width=1.15pt](M)--(Y);
\end{tikzpicture}}}
\end{center}
\caption{(a): GCM with a confounder $C$ for the causal effect of $A$ on $Y$. (b): GCM with one direct and one indirect causal path from $A$ to $Y$. (c): GCM with a confounder $C$ for the effect of $M$ on $Y$.}
\label{fig:GCM}
\vspace{-0.0cm}
\end{figure}
In order to achieve path-specific counterfactual fairness, a decision system needs to be able to discern the causal effect of the sensitive attribute on the decision along the fair and unfair pathways, 
and to disregard the effect along the latter pathways.
\citet{kilbertus17avoiding} and \citet{nabi18fair} 
propose to constrain the learning of the model parameters such that the unfair effect is eliminated or reduced.
However, this approach has several limitations and restrictions: \\[-17pt]
\begin{itemize}[leftmargin=*]
\item It requires specification of the constraints. \citet{nabi18fair} explicitly compute an approximation of the unfair effect, and perform optimization of the model parameters under that constraint that the effect must lie in a small range.
Instead, \citet{kilbertus17avoiding} directly identify a set of constraints on the conditional distribution of the decision variable that eliminate the unfair effect. Both approaches share the limitation of relying on linear relations among the random variables of the model.
Furthermore, imposing constraints on the model parameters distorts the underlying data-generation process.\\[-10pt]
\item To take a decision, the system in \citet{nabi18fair} requires averaging over all variables that are descendants of the sensitive attribute through unfair pathways.
This can negatively impact the system's predictive accuracy. 
\citet{kilbertus17avoiding} also unnecessarily remove information from a subset of such descendants.\\[-14pt]
\end{itemize}
We propose a different approach in which, rather than imposing constraints on the model parameters to eliminate unfair effects, 
the system takes a fair decision by \emph{correcting} the variables that are descendants of the sensitive attribute through unfair pathways. 
The correction aims at eliminating the unfair information contained in the descendants, induced by the sensitive attribute, while retaining the remaining fair information.
This approach more naturally achieves path-specific counterfactual fairness without completely disregarding information from the problematic descendants.
By leveraging recent developments in deep learning and approximate inference, we are able to propose a method that is widely applicable to complex, non-linear scenarios.

\section{Background on Causality}
Causal relationships among random variables can be visually expressed using Graphical Causal Models (GCMs), which are a special case of Graphical Models (see \citet{chiappa14explicit} for a quick introduction on GMs). 
GCMs comprise nodes representing the variables, and links describing statistical and causal relationships among them. In this work, we will restrict ourselves to directed acyclic graphs, 
\ie~graphs in which a node cannot be an \emph{ancestor} of itself. In directed acyclic graphs, the joint distribution over all nodes is given by the product of the conditional distributions of each node $X_i$ given its \emph{parents} $\textrm{par}(X_i)$,  $p(X_i|\textrm{par}(X_i))$. 

GCMs enable us to give a graphical definition of causes and causal effects: if a variable $Y$ is a \emph{child} of another variable $A$, then $A$ is a \emph{direct cause} of $Y$. 
If $Y$ is a \emph{descendant} of $A$, then $A$ is a \emph{potential cause} of $Y$. 

It follows that the causal effect of $A$ on $Y$ can be seen as the information that $A$ sends to $Y$ through \emph{causal paths}, \ie~directed paths,
or as the conditional distribution of $Y$ given $A$ restricted to those paths. This implies that the causal effect of $A$ on $Y$ coincides with $p(Y|A)$ only if 
there are no open noncausal, \ie~undirected, paths between $A$ and $Y$. An example of an open undirected path from $A$ to $Y$ is given by $A\leftarrow C \rightarrow Y$ in \figref{fig:GCM}(a): 
the variable $C$ is said to be a \emph{confounder} for the effect of $A$ on $Y$. 

If confounders are present, then the causal effect can be retrieved by \emph{intervening} on $A$, 
which replaces the conditional distribution of $A$ with, in the case considered in this paper, a fixed value $a$. 
For the model in \figref{fig:GCM}(a), intervening on $A$ by setting it to the fixed value $a$ would correspond to replacing $p(A|C)$ with a delta distribution $\delta_{A=a}$, thereby removing the link from $C$ to $A$ and leaving the remaining conditional distributions $p(Y|A,C)$ and $p(C)$ unaltered.
After intervention, the causal effect is given by the conditional distribution in the intervened graph, namely by $p^*(Y|A=a) = \sum_C p(Y|A=a,C)p(C)$.

We define $Y_{A=a}$ to be the random variable that results from such constrained conditioning, \ie~with distribution $p(Y_{A=a}) = p^*(Y|A=a)$. $Y_{A=a}$ is called a \emph{potential outcome} variable and, when not ambiguous, we will refer to it with the shorthand $Y_a$.

\section{Causal Effect along Different Pathways}
We are interested in separating the causal effect of a sensitive attribute $A$ on the decision variable $Y$ into the contribution along fair and unfair causal pathways.
For simplicity, we assume that $A$ can only take two values $a$ and $a'$, and that $a'$ is a baseline value (these assumptions can readily be relaxed). For example, in the Berkeley alleged sex bias case, 
$a$ and $a'$ would correspond to female and male applicants respectively. 
Under these assumptions, path-specific counterfactual fairness is achieved when the difference between the causal effects along the unfair pathways for $A=a$ and $A=a'$ is small. 
In the remainder of the paper, we will refer to (the average of) this difference as the path-specific effect (PSE). 

\citet{nabi18fair} suggest explicitly computing the PSE and constraining it to lie in a small range during learning of the model parameters. 
\citet{pearl01direct,pearl12causal} and \citet{shpitser13counterfactual} indicate how and when the PSE can be expressed through integrations over conditional distributions involving only observed variables. 
However, except for linear/low dimensional scenarios, the actual computation of the PSE can be non-trivial, due to the intractability of the integrations. 
This issue is addressed with approximations and linearity assumptions. 
Another disadvantage of this approach is that, to take a decision, the system requires averaging over all variables that are descendants of $A$ through unfair pathways, losing their predictive power.

To avoid these difficulties and limitations, we instead propose to implicitly remove the PSE from the decision by intervening on $A$, setting it to the baseline value along the unfair pathways.
This correction procedure enables us to remove the PSE without explicitly computing it and to retain the fair information contained in the problematic descendants of $A$.

Before introducing our approach, for completeness we summarize the method outlined in \citet{pearl01direct,pearl12causal} and \citet{shpitser13counterfactual} for the explicit computation of the PSE.
Readers already familiar with this topic can skip to \secref{sec:ArtificiaDataExample}.

\subsection{Direct and Indirect Effect}
The simplest scenarios for the identification of the effect of $A$ on $Y$ along a subset of pathways are the identification along the direct path $A\rightarrow Y$ (\emph{direct effect}) and along the indirect causal paths $A\rightarrow\ldots\rightarrow Y$ (\emph{indirect effect}).

The direct effect can be estimated by computing the difference between 1) the effect when $A=a$ along the direct path $A\rightarrow Y$ and $A=a'$ along the indirect causal paths $A\rightarrow\ldots\rightarrow Y$
and 2) the effect when $A=a'$ along all causal paths. 

Similarly, the indirect effect can be estimated by computing the difference between 1) the effect when $A=a$ along all causal paths and 2) the effect when $A=a$ along the direct path $A\rightarrow Y$ and $A=a'$ along the indirect causal paths $A\rightarrow\ldots\rightarrow Y$.

Suppose that the GCM contains only one indirect path through a variable $M$, as in \figref{fig:GCM}(b).
We define $Y_{a}(M(a'))$ to be the \emph{counterfactual} random variable that results from the intervention $A=a$ along $A\rightarrow Y$ and the intervention $A=a'$ along $A\rightarrow M\rightarrow Y$. 
The \emph{average direct effect} (ADE) and the \emph{average indirect effect} (AIE) are given by\footnote{In this paper, we consider the \emph{natural} effect, which generally differs from the \emph{controlled} effect; the latter corresponds to intervening on $M$.}
\begin{align*}
\textrm{ADE: } & \av{Y_{a}(M(a'))}-\av{Y_{a'}}, & \textrm{AIE: } \av{Y_{a}} - \av{Y_{a}(M(a'))}\,,
\end{align*}
where, \eg, $\av{Y_{a}}=\int_{Y_{a}} Y_{a}p(Y_{a})$. 
Under the independence assumption $Y_{a,m}\ci M_{a'}$ (\emph{sequential ignorability}), the counterfactual distribution $p(Y_{a}(M(a')))$ can be estimated from non-counterfactual 
distributions, \ie
\begin{align}
p(Y_{a}(M(a'))) &=\int_m p(Y_a(M(a'))|M_{a'}\!=m)p(M_{a'}\!=m)\nonumber\\
&=\int_mp(Y_{a,m}|M_{a'}=m)p(M_{a'}=m)\nonumber\\
&=\int_mp(Y_{a,m})p(M_{a'}=m)\,,
\label{eq:counter}
\end{align}
where to obtain the second line we have used the \emph{consistency} property \cite{pearl16causal}. 
As there are no confounders, intervening coincides with conditioning, namely $p(Y_{a,m})=p(Y|A=a,M=m)$ and $p(M_{a'})=p(M|A=a')$. 

If the GCM contains a confounder for the effect of either $A$ or $M$ on $Y$, \eg~$C$ in \figref{fig:GCM}(c), 
then $p(Y_{a,m})\neq p(Y|A=a,M=m)$. In this case, by following similar arguments as used in \eqref{eq:counter} but conditioning on $C$ (and therefore assuming $Y_{a,m}\ci M_{a'}|C$), 
we obtain
\begin{align}
p(Y_{a}(M(a')))=\int_{m,c} p(Y| a,m,c)p(m|a',c)p(c)\,.
\label{eq:counter-unobs}
\end{align}
If $C$ is unobserved, then the conditional distributions in \eqref{eq:counter-unobs} cannot be computed with simple methods such as regression. 
In this case, the effect is said to be \emph{non-identifiable}. 
However, these distributions could be learned by maximizing the model likelihood $p(A,M,Y)$ using \eg~latent-variable methods.

\subsection{Path-Specific Effect\label{sec:pse}}
In the more complex scenario in which, rather than computing the direct and indirect effects, 
we want to compute the effect along a specific set of paths, 
we can use the formula for the ADE with the appropriate counterfactual variable.
 
This variable can be derived using the following recursive rule \cite{shpitser13counterfactual}.
Set $X'=Y$. Until $X'=A$, repeat
\begin{enumerate}[leftmargin=*]
\item For each direct cause $X$ of $X'$ along a black arrow:\\
If $X$ is $A$, set $A$ to the baseline $a'$ along $A\rightarrow X'$.\\
If $X$ is not $A$, set $X$ to the value $X(a')$ attained by setting $A$ to the baseline $a'$ along $A\rightarrow X$.\\ [-15pt]
\item For each direct cause $X$ of $X'$ along a green arrow:\\ 
If $X$ is $A$, set $A$ to $a$ along $A\rightarrow X'$.\\
If $X$ is not $A$, set $X$ to the value $\gamma_X$ attained under the effect along the path $A\rightarrow,\ldots, \rightarrow X$. Set $X'=X$.\\ [-15pt]
\end{enumerate}
For example, the required counterfactual variable for the effect along the path $A \rightarrow W \rightarrow Y$ in \figref{fig:PSE}(a) is $Y_{a'}(M(a'),W(a,M(a')))$. 
Indeed, in the first iteration, as $A$ and $M$ are direct causes of $Y$ along black arrows, whilst $W$ is a direct cause of $Y$ along a green arrow,  we obtain $Y_{a'}(M(a'),\gamma_W)$. 
In the second iteration, as $M$ is a direct cause of $W$ along a black arrow, whilst $A$ is a direct cause of $W$ along a green arrow, we obtain $\gamma_W=W(a,M(a'))$.
To compute the counterfactual distribution $p(Y_{a'}(M(a'),W(a,M(a'))))$ from non-counterfactual ones, 
we need to assume $\{Y_{a',m,w},M_{a'} \}\ci W_{a,m}$. This gives
\begin{align*}
&p(Y_{a'}(M(a'),W(a,M(a'))))\\
&=\int_{m,w}p(Y_{a',m,w}|M_{a'}=m)p(M_{a'}=m)p(W_{a,m}=w)\,,
\end{align*}
where $p(Y_{a',m,w}|M_{a'}=m)=p(Y|a',m,w)$, $p(M_{a'})=p(M|a')$, and $p(W_{a,m})=p(W|a,m)$.

For the path $A\rightarrow Y$ in \figref{fig:PSE}(b), we would need instead $p(Y_{a}(M(a'),W(a')))$. 
Under the assumption $Y_{a,m,w} \ci \{M_{a'}, W_{a'} \}$, we would obtain $p(Y_{a}(M(a'),W(a')))=\int_{m,w}p(Y_{a,m,w})p(M_{a'},W_{a'})$.
However, $p(Y_{a,m,w})\neq p(Y|a,m,w)$ and therefore we would need to condition on $C$ as in \eqref{eq:counter-unobs}. 
If $C$ is unobserved, then the effect along this path is not identifiable. In the Appendix, we describe the graphical method introduced in \citet{shpitser13counterfactual} for understanding whether a PSE is identifiable.

\section{Path-Specific Counterfactual Fairness\label{sec:ArtificiaDataExample}}
\begin{figure}[t]
\begin{center}
\subfigure[]{
\scalebox{0.62}{
\begin{tikzpicture}[dgraph]
\node[ocont] (C) at (6,1.5) {$C$};
\node[ocont] (M) at (3.5,1) {$M$};
\node[ocont] (W) at (4.75,0.25) {$W$};
\node[ocont] (X) at (3.5,-0.5) {$A$};
\node[ocont] (Y) at (6,-0.5) {$Y$};
\draw[line width=1.15pt](C)--(M);\draw[line width=1.15pt](C)--(Y);
\draw[line width=1.15pt](X)--(M);\draw[line width=1.15pt, green](X)--(W);\draw[line width=1.15pt](X)--(Y);\draw[line width=1.15pt](M)--(W);\draw[line width=1.15pt, green](W)--(Y);
\draw[line width=1.15pt](M)to [bend right=-35](Y);
\end{tikzpicture}}}
\subfigure[]{
\scalebox{0.62}{
\begin{tikzpicture}[dgraph]
\node[ocont] (C) at (6,1.5) {$C$};
\node[ocont] (M) at (3.5,1) {$M$};
\node[ocont] (W) at (4.75,0.25) {$W$};
\node[ocont] (X) at (3.5,-0.5) {$A$};
\node[ocont] (Y) at (6,-0.5) {$Y$};
\draw[line width=1.15pt](C)--(M);\draw[line width=1.15pt](C)--(Y);
\draw[line width=1.15pt](X)--(M);\draw[line width=1.15pt](X)--(W);\draw[line width=1.15pt,green](X)--(Y);\draw[line width=1.15pt](M)--(W);\draw[line width=1.15pt](W)--(Y);
\draw[line width=1.15pt](M)to [bend right=-35](Y);
\end{tikzpicture}}}
\subfigure[]{
\scalebox{0.66}{
\begin{tikzpicture}[dgraph]
\node[ocont] (C) at (3,2.5) {$C$};
\node[ocont] (A) at (0,1) {$A$};
\node[ocont] (M) at (1.5,1) {$M$};
\node[ocont] (L) at (3,1) {$L$};
\node[ocont] (Y) at (4.5,1) {$Y$};
\draw[line width=1.15pt,green](A)--(M);\draw[line width=1.15pt, postaction={draw,green,dash pattern= on 3pt off 6pt,dash phase=4pt}][line width=1.15pt, black,dash pattern= on 3pt off 6pt] (M)--(L);
\draw[line width=1.15pt, postaction={draw,green,dash pattern= on 3pt off 6pt,dash phase=4pt}][line width=1.15pt, black,dash pattern= on 3pt off 6pt] (L)--(Y);
\draw[line width=1.15pt, postaction={draw,green,dash pattern= on 3pt off 6pt,dash phase=4pt}][line width=1.15pt, black,dash pattern= on 3pt off 6pt] (M)to [bend left=-30](Y);
\draw[line width=1.15pt](A)to [bend left=-25](L);
\draw[line width=1.15pt,green](A)to [bend right=-25](Y);
\draw[line width=1.15pt](C)--(M);\draw[line width=1.15pt](C)--(L);\draw[line width=1.15pt](C)--(Y);
\end{tikzpicture}}}
\caption{(a)-(b): GCMs in which we are interested in the effects along the green paths. (c): GCM corresponding to \eqref{eq:lm}.}
\label{fig:PSE}
\end{center}
\end{figure}
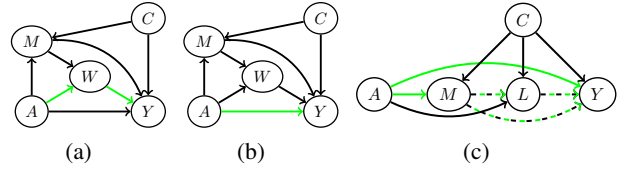

To gain insights into the problem of path-specific fairness, consider the following linear model
\begin{align}
&A=\textrm{Bernoulli}(\pi), \hskip0.1cm C = \epsilon_c\,,\nonumber\\
&M=\theta^m+\theta^m_{a}A+\theta^m_{c}C+\epsilon_m\,,\nonumber\\
&L=\theta^l+\theta^l_{a}A+\theta^l_{c}C+\theta^l_{m}M+\epsilon_l\,,\nonumber\\
&Y=\theta^y+\theta^y_{a}A+\theta^y_{c}C+\theta^y_{m}M+\theta^y_{l}L+\epsilon_y\,.
\label{eq:lm}
\end{align}
The variables $A, C, M, L$ and $Y$ are observed, whilst $\epsilon_a$, $\epsilon_c$, $\epsilon_m$ and $\epsilon_l$ are unobserved independent zero-mean Gaussian terms with variance $\sigma^2_a,\sigma^2_c,\sigma^2_m,\sigma^2_l$ and $\sigma^2_y$.
The GCM corresponding to this model is depicted in \figref{fig:PSE}(c).

We want to learn to predict $Y$ from $A, C, M$ and $L$. However, $A$ is a sensitive attribute, and its direct effect on $Y$ and effect through $M$ is considered unfair.
Therefore, to obtain a fair decision system, we need to disregard the PSE of $A$ on $Y$ along the direct path $A\rightarrow Y$ and the paths passing through $M$, $A\rightarrow M \rightarrow,\ldots,\rightarrow Y$, namely along the green and dashed green-black links of \figref{fig:PSE}(c). 
Notice that the dashed green-black links differ fundamentally from the green links; they contain unfairness only as a consequence of $A \rightarrow M$, corresponding to the parameter $\theta^m_{a}$, being unfair.

Assume $a'=0$ is the baseline value of $A$. 
Using the recursive rule described in \secref{sec:pse}, we can deduce that the counterfactual variable required to estimate the desired PSE is $Y_a(M(a), L(a', M(a)))$ and has distribution
\begin{align*}
\int_{C,M,L} p(Y|a,C,M,L)p(L|a',C,M)p(M|a,C)p(C)\,.
\end{align*}
The mean of this distribution is $\theta^y+\theta^y_{m}\theta^m+\theta^y_l(\theta_l+\theta^l_m\theta^m)+\theta^y_{a}a+\theta^y_{m}\theta^m_{a}a+\theta^y_{l}(\theta^l_{a}a'+\theta^l_{m}\theta^m_{a}a)$.
The PSE, namely difference between this quantity and the mean of the effect of $A=a'$, is therefore given by $\textrm{PSE}=\theta^y_{a}(a-a')+\theta^y_{m}\theta^m_{a}(a-a')+\theta^y_{l}\theta^l_{m}\theta^m_{a}(a-a')$ 
which, as $a'=0$ and $a=1$, simplifies to 
\begin{align*}
\textrm{PSE}=\theta^y_{a}+\theta^m_{a}(\theta^y_{m}+\theta^y_{l}\theta^l_{m})\,.
\end{align*}

\citet{nabi18fair} suggest learning the subset of parameters $\theta=\{\theta^m,\theta^m_a,\theta^m_c,\theta^l,\theta^l_a,\theta^l_c,\theta^l_m,\theta^y,\theta^y_a,\theta^y_c, \theta^y_m,\theta^y_l\}$ 
whilst constraining the PSE to lie in a small range. They then form a prediction $\hat y^n$ of a new instance $\{a^n,c^n,m^n,l^n\}$ as the mean of 
$p(Y|a^n, c^n)=\int_{M,L}p(Y|a^n,c^n,M,L)p(L|a^n,c^n,M)p(M|a^n,c^n)$, \ie~by integrating out $M$ and $L$. Therefore, to form $\hat y^n$, the individual-specific information within $m^n$ and $l^n$ is disregarded. 
With $\Theta^m=\theta^m+{\myred\theta^m_{a}}+\theta^m_{c}c^n$ and $a^n=1$, 
$\hat y^n=\theta^y+{\myred \theta^y_{a}}+\theta^y_{c}c^n+\theta^y_{m}\Theta^m+\theta^y_{l}(\theta^l+{\myred \theta^l_{a}}+\theta^l_{c}c^n+\theta^l_{m}\Theta^m)$. If $a^n=0$ the terms in red are omitted. 

\citet{nabi18fair} justify averaging over $M$ and $L$ due to the need to account for the constraints that are potentially imposed on $\theta^m_{a}$ and $\theta^l_{m}$, 
and suggest that this is an inherent limitation of path-specific fairness. It is true that, if a constraint is imposed on a parameter, the corresponding variable needs to be integrated out 
to ensure that such a constraint is taken into account in the prediction. However, all the descendants of $A$ along unfair pathways must be integrated out, regardless of whether or not constraints are imposed on the corresponding parameters. 
Indeed the observations $m^n$ and $l^n$ contain the unfair part of the effect from $A$, which needs to be disregarded. 
To better understand this point, notice that we could obtain 
$\textrm{PSE}=\theta^y_{a}+\theta^m_{a}(\theta^y_{m}+\theta^y_{l}\theta^l_{m})=0$ by a priori imposing the constraints $\theta^y_{a}=0$ and $\theta^m_{a}=0$,
which would not constrain $\theta^l_{m}$. However, to form a prediction $\hat y^n$, we would still need to integrate over $L$, as the observation $l^n$ contains the problematic term $\theta^y_{l}\theta^l_{m}\theta^m_{a}$.

In this simple case, we could avoid having to integrate over $M$ and $L$ by a priori imposing the constraints $\theta^y_{a}=0$ and $\theta^y_{m}=-\theta^y_{l}\theta^l_{m}$, \ie~constraining 
the conditional distribution used to form the prediction $\hat y^n$, $p(Y|A,C,M,L)$. This coincides with the constraint proposed in \citet{kilbertus17avoiding} to avoid proxy discrimination. 
However, this approach achieves removal of the problematic terms in $m^n$ and $l^n$ by canceling out the entire $m^n$ from the prediction. 
This is also suboptimal, as all information within $m^n$ is disregarded. Furthermore, it is not clear how to extend this approach to more complex scenarios.

Our insight is to notice that we can achieve a fair prediction of a new instance $\{a^n=1,c^n,m^n,l^n\}$ whilst retaining all fair information using $\hat y^n_{\textrm{fair}} =\av{Y}_{p(Y|a^n,c^n,m^n,l^n)} - \textrm{PSE}$
\begin{align*}
\hat y^n_{\textrm{fair}} &=\theta^y+\theta^y_{c}c^n+\theta^y_{m}m^n+\theta^y_{l}l^n -\theta^m_{a}(\theta^y_{m}+\theta^y_{l}\theta^l_{m})\,.
\end{align*}
Alternatively, the same solution can be achieved as follows. We first estimate the values of the noise terms $\epsilon^n_m$ and $\epsilon^n_l$ from $a^n, c^n, m^n, l^n$ and the model equations, \ie~$\epsilon^n_m = m^n - \theta^m - \theta^m_{a} - \theta^m_{c}c^n$ and 
$\epsilon^n_l = \theta^l - \theta^l_{a} - \theta^l_{c}c^n - \theta^l_{m}m^n$. We then obtain fair transformations of $m^n$ and $l^n$ and a fair prediction $\hat y^n_{\textrm{fair}}$
by substituting $\epsilon^n_m$ and $\epsilon^n_l$ into the model equations with the problematic terms $\theta^m_{a}$ and $\theta^l_{a}$ removed,
\begin{align}
&m^n_{\textrm{fair}} = \theta^m + \cancel{\theta^m_{a}} + \theta^m_{c}c^n  + \epsilon^n_m\,,\nonumber\\
&l^n_{\textrm{fair}}  = \theta^l + \theta^l_{a} + \theta^l_{c}c^n + \theta^l_{m}m^n_{\textrm{fair}} +\epsilon^n_l\,,\nonumber\\
&\hat y^n_{\textrm{fair}}  =\theta^y+ \cancel{\theta^y_{a}} +\theta^y_{c}c^n+\theta^y_{m}m^n_{\textrm{fair}}+\theta^y_{l}l^n_{\textrm{fair}}\nonumber\\
&\hskip0.5cm=\theta^y+\theta^y_{a} -\theta^y_{a} + \theta^y_{c}c^n+ \theta^y_{m}(m^n-\theta^m_{a})\nonumber\\
&\hskip0.5cm+\theta^y_{l}(l^n - \theta^l_{m}\theta^m_{a})\,.
\label{eq:CR}
\end{align}
In other words, we compute a fair prediction by intervening on $A$, setting it to the baseline value along the links that create unfairness $A\rightarrow M$ and $A\rightarrow Y$.
This is effectively an extension of the procedure for performing counterfactual reasoning in structural equation models \cite{pearl00causality}, where the counterfactual correction is only restricted to the problematic links $A\rightarrow M$ and $A\rightarrow Y$.
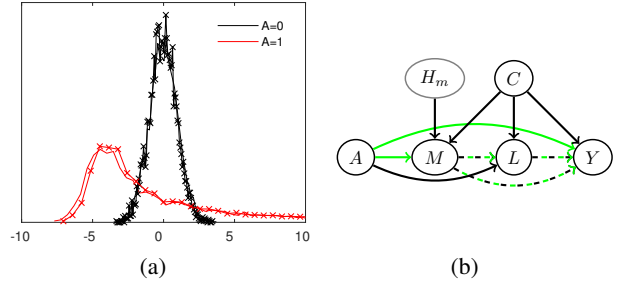
\begin{figure}[t]
\begin{center}
\subfigure[]{
\includegraphics[height=3.2cm,width=4.0cm]{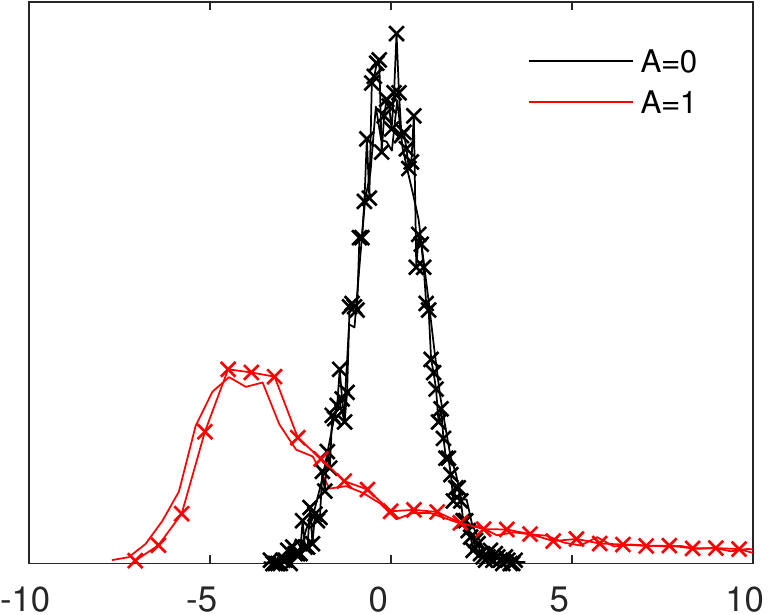}}
\hskip0.1cm
\subfigure[]{
\scalebox{0.70}{
\begin{tikzpicture}[dgraph]
\node[hcont] (H) at (1.5,2.5) {$H_m$};
\node[ocont] (C) at (3,2.5) {$C$};
\node[ocont] (A) at (0,1) {$A$};
\node[ocont] (M) at (1.5,1) {$M$};
\node[ocont] (L) at (3,1) {$L$};
\node[ocont] (Y) at (4.5,1) {$Y$};
\node[] at (0,-0.5) {$$};
\draw[line width=1.15pt,green](A)--(M);\draw[line width=1.15pt, postaction={draw,green,dash pattern= on 3pt off 6pt,dash phase=4pt}][line width=1.15pt, black,dash pattern= on 3pt off 6pt] (M)--(L);
\draw[line width=1.15pt, postaction={draw,green,dash pattern= on 3pt off 6pt,dash phase=4pt}][line width=1.15pt, black,dash pattern= on 3pt off 6pt] (L)--(Y);
\draw[line width=1.15pt, postaction={draw,green,dash pattern= on 3pt off 6pt,dash phase=4pt}][line width=1.15pt, black,dash pattern= on 3pt off 6pt] (M)to [bend left=-30](Y);
\draw[line width=1.15pt](A)to [bend left=-25](L);
\draw[line width=1.15pt,green](A)to [bend right=-25](Y);
\draw[line width=1.15pt](C)--(M);\draw[line width=1.15pt](C)--(L);\draw[line width=1.15pt](C)--(Y);
\draw[line width=1.15pt](H)--(M);
\end{tikzpicture}}}
\end{center}
\caption{(a): Empirical distribution of $\epsilon^n_m$ for the case in which $m^n$ is generated by \eqref{eq:lm} with an extra non-linear term $f(A,C)$ (continuous lines). Histograms of $\tilde p(H_m|A)$ (crossed lines), see (b). 
(b): Modification of the GCM corresponding to \eqref{eq:lm} to include an explicit latent variable $H_m$ for the generation of $M$. }
\label{fig:LM}
\end{figure}
\paragraph{Model-Observations Mismatch.}  
Whilst this approach provides us with an elegant and straightforward way to impose path-specific counterfactual fairness, 
if there is a mismatch between the data-generation processes assumed by the learned model and underlying the observations, fairness is most likely not achieved. 

Consider, for example, the case in which we assume the data-generation process of \eqref{eq:lm}, but the observed $m^n$, $n=1,\ldots,N$, are generated from a modified version of \eqref{eq:lm} containing an extra non-linear term $f(A,C)$. 
The learned $\theta$ would not be able to describe this non-linear term, which would therefore be absorbed into the noise values $\epsilon^n_m$, making the noise and $A$ dependent, as shown in \figref{fig:LM}(a). 

To solve this issue, we propose to introduce an explicit latent variable $H_m$ for the generation of $M$, \ie~$M=\theta^m+\theta^m_{a}A+\theta^m_{c}C+H_m+\epsilon_m$,
obtaining the GCM of \figref{fig:LM}(b). Define
\begin{align*}
\tilde p(H_m|A=a) = \frac{1}{N_a}\sum_{n=1}^{N_a} p(H_m|a^n=a,c^n,m^n,l^n)\,,
\end{align*}
where $N_a$ indicates the number of observations for which $a^n=a$.
We encourage $\tilde p(H_m|A=a)$ to have small dependence on $A$ during training through the maximum mean discrepancy (MMD) criterion \cite{gretton12kernel}.
We can then use, \eg, the mean of $p(H_m|a^n,c^n,m^n,l^n)$, rather than $\epsilon^n_m$, in \eqref{eq:CR}. ($p(H_m|a^n,c^n,m^n,l^n)$ is Gaussian and can be computed analytically, see the Appendix).

\citet{kusner17counterfactual}, who also use a latent-variable approach, do not enforce small dependence. 
To demonstrate that this is necessary, we learned the parameters of the modified model with Gaussian distribution $p(H_m)$, using an expectation maximization approach. 
$\tilde p(H_m|A)$ is shown in \figref{fig:LM}(a). As we can see, the extra term $f(A,C)$ is absorbed by the latent variable. In other words, even if $p(H_m|A)=p(H_m)$, the mismatch between the model and the observations implies $\tilde p(H_m|A) \neq \tilde p(H_m)$.

In the next section, we explain how the approach described above can be implemented in a general algorithm that is applicable to complex, non-linear models.

\subsection{Latent Inference-Projection Approach\label{sec:lva}}
For addressing a more general data-generation process mismatch than the one considered above, we need to explicitly incorporate a latent variable for each descendant of the sensitive attribute that needs to be corrected. 
General equations for the GCM of \figref{fig:PSE}(c) with extra latent variables $H_m$ and $H_l$ are
\begin{align*}
&A\sim p_{\theta}(A), C \sim p_{\theta}(C)\,,\nonumber\\
&M \sim p_{\theta}(M|A,C,H_m), \hskip0.1cm L \sim p_{\theta}(L|A,C,M,H_l)\,,\\
&Y \sim p_{\theta}(Y|A,C,M,L)\,.
\end{align*}
If $M$ is categorical, $p_{\theta}(M|A,C,H_m)=f_{\theta}(A,C,H_m)$, where $f_{\theta}(A,C,H_m)$ can be any function, \eg~a neural network.
If $M$ is continuous, $p_{\theta}(M|A,C,H_m)$ is Gaussian with mean $f_{\theta}(A,C,H_m)$. 

The model likelihood $p_{\theta}(A,C,M,L,Y)$, and posterior distributions $p_{\theta}(H_m|A,C,M,L)$ and $p_{\theta}(H_l|A,C,M,L)$ required to form fair predictions, are generally intractable. 
We address this with a variational approach that computes Gaussian approximations $q_{\phi}(H_m|A,C,L,M)$ and $q_{\phi}(H_l|A,C,L,M)$, of $p_{\theta}(H_m|A,C,M,L)$ and $p_{\theta}(H_l|A,C,M,L)$ respectively, 
parametrized by $\phi$, as discussed in detail below. 

After learning $\theta$ and $\phi$, to form a fair prediction $\hat y^n_{\textrm{fair}}$ of a new instance $\{a^n=a,c^n,m^n,l^n\}$, we proceed analogously to \eqref{eq:CR} using a Monte-Carlo approach: 
we first draw samples $h^{n,i}_m \sim q_{\phi}(H_m|a^n,c^n,m^n,l^n)$ and $h^{n,i}_l \sim q_{\phi}(H_l|a^n,c^n,m^n,l^n)$, for $i=1,\ldots,I$, and then form 
\begin{align}
&m^{n,i}_{\textrm{fair}} \sim  p_{\theta}(M|a',c^n,h^{n,i}_m)\,,\nonumber\\
&l^{n,i}_{\textrm{fair}} \sim p_{\theta}(L|a^n,c^n,m^{n,i}_{\textrm{fair}},h^{n,i}_l)\,,\nonumber\\
&\hat y^n_{\textrm{fair}}=\frac{1}{I}\sum_{i=1}^I\av{Y}_{p_{\theta}(Y|a',c^n,m^{n,i}_{\textrm{fair}},l^{n,i}_{\textrm{fair}})}\,.
\label{eq:gm}
\end{align} 
If we group the observed and latent variables as $V=\{A,C,M,L,Y\}$ and $H=\{H_m,H_l\}$ respectively,
the variational approximation $q_{\phi}(H|V)$ to the intractable posterior $p_{\theta}(H|V)$ is obtained
by finding the variational parameters $\phi$ that minimize the Kullback-Leibler divergence  $KL(q_{\phi}(H|V)||p_{\theta}(H|V))$. 
This is equivalent to maximizing a lower bound ${\cal F}_{\theta,\phi}$ on the log marginal likelihood 
$\log p_{\theta}(V)\geq {\cal F}_{\theta,\phi}$ with
\begin{align*}
{\cal F}_{\theta,\phi} =  -\av{\log q(H|V)}_{q(H|V)}+\av{\log p(V,H)}_{q(H|V)}\,.
\end{align*}
In our case, rather than $q(H|V)$, we use $q(H|V^* \equiv V\setminus Y)$. 
Our approach is therefore to learn simultaneously the latent embedding and predictive distributions in \eqref{eq:gm}. 
This could be preferable to other causal latent variable approaches such as the $\textrm{FairLearning}$ algorithm proposed in \citet{kusner17counterfactual},
which separately learns a predictor of $Y$ using samples from the previously inferred latent variables and from the non-descendants of $A$.

In order for ${\cal F}_{\theta,\phi}$ to be tractable conjugacy is required, which heavily restricts the family of models that can be used. 
This issue can be addressed with a Monte-Carlo approximation recently introduced in \citet{kingma14autoencoding} and \citet{rezende14stochastic}.
This approach represents $H$ as a non-linear transformation $H=f_{\phi}(\mathcal{E})$ of a random variable $\mathcal{E}$ from a parameter free distribution $q_{\epsilon}$.
As we choose $q$ to be Gaussian, $H=\mu_{\phi}+\sigma_{\phi}\mathcal{E}$ with $q_{\epsilon}={\cal N}(0,1)$ for the univariate case.
This enables us to rewrite the bound as
\begin{align*}
{\cal F_{\theta,\phi}}&=-\av{\log q(H\!=\!f_{\phi}(\mathcal{E}))+\log p(V,H\!=\!f_{\phi}(\mathcal{E}))}_{q_{\epsilon}}\,.
\end{align*}
The first part of the gradient of ${\cal F_{\theta,\phi}}$ with respect to $\phi$, $\nabla_{\phi}{\cal F_{\theta,\phi}}$,  can be computed analytically, whilst the second part is approximated by
\begin{align*}
\av{\nabla_{\phi} \log p(V,&H=f_{\phi}(\mathcal{E}))}_{q_{\epsilon}}\\ 
&\approx  \frac{1}{I}\sum_{i=1}^I \nabla_{\phi} \log p(V,h^i=f_{\phi}(\epsilon^i)) , \hskip0.05cm \epsilon^i\sim q_{\epsilon}\,.
\end{align*}
The variational parameters $\phi$ are parametrized by a neural network taking as input $V^*$. 
In order to ensure that $\tilde q(H|A)$ does not depend on $A$, we follow a MMD penalization approach as in \citet{louizos16fair}.
This approach adds a MMD estimator term, approximated using random Fourier features, to the bound ${\cal F_{\theta,\phi}}$, weighted by a factor $\beta$.

\section{Experiments}
In this section, we first show that our approach can disregard unfair information without a big loss in accuracy on a biased version of the Berkeley Admission dataset. 
We then test our method on two datasets previously considered in the causal and fairness literature, namely the UCI Adult and German Credit datasets. 
Experimental details are given in the Appendix.

\subsection{The Berkeley Admission Dataset\label{sec:BerkeleyDataset}} 
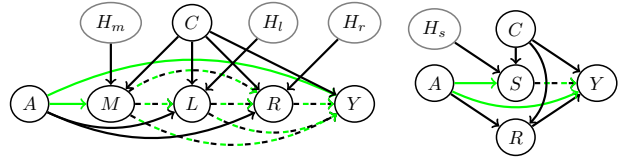
\begin{figure}[t]
\begin{center}
\scalebox{0.72}{
\begin{tikzpicture}[dgraph]
\node[hcont] (Hm) at (1.5,2.5) {$H_m$};
\node[ocont] (C) at (3,2.5) {$C$};
\node[hcont] (Hl) at (4.5,2.5) {$H_l$};
\node[hcont] (Hr) at (6,2.5) {$H_r$};
\node[ocont] (A) at (0,1) {$A$};
\node[ocont] (M) at (1.5,1) {$M$};
\node[ocont] (L) at (3,1) {$L$};
\node[ocont] (R) at (4.5,1) {$R$};
\node[ocont] (Y) at (6,1) {$Y$};
\draw[line width=1.15pt,green](A)--(M);
\draw[line width=1.15pt, postaction={draw,green,dash pattern= on 3pt off 6pt,dash phase=4pt}][line width=1.15pt, black,dash pattern= on 3pt off 6pt] (M)--(L);
\draw[line width=1.15pt, postaction={draw,green,dash pattern= on 3pt off 6pt,dash phase=4pt}][line width=1.15pt, black,dash pattern= on 3pt off 6pt] (L)--(R);
\draw[line width=1.15pt, postaction={draw,green,dash pattern= on 3pt off 6pt,dash phase=4pt}][line width=1.15pt, black,dash pattern= on 3pt off 6pt] (R)--(Y);
\draw[line width=1.15pt, postaction={draw,green,dash pattern= on 3pt off 6pt,dash phase=4pt}][line width=1.15pt, black,dash pattern= on 3pt off 6pt] (M)to [bend right=-35](R);
\draw[line width=1.15pt, postaction={draw,green,dash pattern= on 3pt off 6pt,dash phase=4pt}][line width=1.15pt, black,dash pattern= on 3pt off 6pt] (M)to [bend left=-30](Y);
\draw[line width=1.15pt, postaction={draw,green,dash pattern= on 3pt off 6pt,dash phase=4pt}][line width=1.15pt, black,dash pattern= on 3pt off 6pt] (L)to [bend left=-30](Y);
\draw[line width=1.15pt](A)to [bend left=-25](L);\draw[line width=1.15pt](A)to [bend left=-25](R);\draw[line width=1.15pt,green](A)to [bend right=-25](Y);
\draw[line width=1.15pt](C)--(M);\draw[line width=1.15pt](C)--(L);\draw[line width=1.15pt](C)--(R);\draw[line width=1.15pt](C)--(Y);
\draw[line width=1.15pt](Hl)--(L);\draw[line width=1.15pt](Hm)--(M);\draw[line width=1.15pt](Hr)--(R);
\end{tikzpicture}}
\hskip0.2cm
\scalebox{0.72}{
\begin{tikzpicture}[dgraph]
\node[ocont] (S) at (1.5,1) {$S$};
\node[ocont] (A) at (0,1) {$A$};
\node[ocont] (C) at (1.5,2) {$C$};
\node[ocont] (R) at (1.5,0) {$R$};
\node[hcont] (H) at (0,2) {$H_s$};
\node[ocont] (Y) at (3,1) {$Y$};
\draw[line width=1.15pt,green](A)--(S);
\draw[line width=1.15pt](H)--(S);
\draw[line width=1.15pt](C)--(S);\draw[line width=1.15pt](C)--(Y);
\draw[line width=1.15pt](A)--(R);\draw[line width=1.15pt](R)--(Y);
\draw[line width=1.15pt, postaction={draw,green,dash pattern= on 3pt off 6pt,dash phase=4pt}][line width=1.15pt, black,dash pattern= on 3pt off 6pt] (S)--(Y);
\draw[line width=1.15pt,green](A)to [bend right=25](Y);
\draw[line width=1.15pt](C)to [bend right=-45](R);
\end{tikzpicture}}
\end{center}
\caption{(a): GCM for the UCI Adult dataset. (b): GCM for the UCI German Credit dataset.}
\label{fig:Adult}
\end{figure}

In order to provide a test-case for our methodology, we consider the Berkeley Admission dataset, which contains sex $A$, department choice $D$, and admission decision $Y$ for 4,526 applicants (all variables are categorical).
Counts indicate that 45.5\% of male applicants are admitted, versus only 30.4\% of female applicants, 
\ie~$p(Y|A=0) \neq p(Y|A=1)$. This discrepancy would appear to indicate gender bias. However, the potential outcome variable $Y_D$, resulting from intervening on $D$, does not depend on $A$.
Therefore the dependence of $Y$ on $A$ is entirely through department choice $D$: the reason being that women are applying to departments with lower admission rates.

We modified the dataset to favor male applicants and to discriminate against female applicants, \ie~we added a direct path $A\rightarrow Y$ making $Y_D$  dependent on $A$.
On this new dataset, when predicting using only the link $D\rightarrow Y$ through $p(Y_D)$ we obtained 67.9\% accuracy, whilst when predicting using both $D\rightarrow Y$
and $A\rightarrow Y$ through $p(Y|A,D)$ ($=p(Y_D|A)$) we obtained 71.6\% accuracy, indicating that the link $A\rightarrow Y$ has a strong impact on the decision.

We used the latent inference-projection approach described in \secref{sec:lva} to disregard the effect along the direct path $A\rightarrow Y$.
Whilst $D$ does not require correction, we nevertheless used a latent variable $H_d$ 
to test how much information about $D$ we lose by projecting into a latent space and back. As the bias along the path $A\rightarrow Y$ 
affects both male and female applicants,  we could not use either of the two sexes as a baseline -- instead, we averaged over values of $A$ sampled from $p(A)$.

We divided the dataset into training and test sets of sizes 3,500 and 1,026 respectively, and used $I=1000$ for the Monte-Carlo approximation in \eqref{eq:gm}.
Our method gave a test accuracy of 67.1\%, which is close to the desired 67.9\%, indicating that we are disregarding the unfair information
whilst only losing a small amount of information through latent inference-projection.

\subsection{The UCI Adult Dataset\label{sec:AdultDataset}}
The Adult dataset from the UCI repository \cite{lichman13uci} contains 14 attributes including age, working class, education level, marital status, occupation,
relationship, race, gender, capital gain and loss, working hours, and nationality for 48,842 individuals; 32,561 and 16,281 for the training and test sets respectively. 
The goal is to predict whether the individual's annual income is above or below \$50,000. 
We assume the GCM of \figref{fig:Adult}(a), where $A$ corresponds to the protected attribute sex, $C$ to the duple age and nationality, 
$M$ to marital status, $L$ to level of education, $R$ to the triple working class, occupation, and hours per week, and $Y$ to the income class\footnote{We omit race, and capital gain and loss (although including capital gain and loss would increase test accuracy from 82.9\% to 84.7\%) to use the same attributes as in \citet{nabi18fair}.}.
Age, level of education and hours per week are continuous, whilst sex, nationality, marital status, working class, occupation, and income are categorical.
Besides the direct effect $A \rightarrow Y$, we would like to remove the effect of $A$ on $Y$ through marital
status, namely along the paths $A\rightarrow M \rightarrow,\ldots,\rightarrow Y$. 
This GCM is similar to the one analyzed in \secref{sec:pse} and, except for the latent variables, is the same as the one used in \citet{nabi18fair}.

\begin{table}[]
\caption{In order columns represent: unfair test accuracy, fair test accuracy, and MMD values for $H_m$, $H_l$, and $H_r$  ($\times$ 10,000) for the UCI Adult dataset. Rows represent values after 5,000, 8,000, 15,000, and 20,000 training steps.}
\label{tab:Adult}
\vskip0.25cm
\begin{center}
\begin{small}
\begin{tabular}{ccccccc}
\toprule
82.88\%  & 81.66\%  & 610.85  & 13.31  & 3.73 & 3.10 & 3.12\\
82.85\%  &  80.21\%  & 6.73 & 2.80  & 3.75  &   2.88  &3.10\\
82.71\%  &  79.41\%   & 2.97  & 3.45 & 0.25  &  0.07 & 0.49\\
80.60\%  &  73.98\%  &  3.19   & 6.31  & 0.22 &   0.10 & 0.47\\
\bottomrule
\end{tabular}
\end{small}
\end{center}
\end{table}

\citet{nabi18fair} assume that all variables, except $A$ and $Y$ are continuous, and linearly related, except $Y$ for which $p(Y=1|\textrm{par}(Y))=\pi=\sigma(\theta^y+\sum_{X_i\in\textrm{par}(Y)}\theta^y_{x_i}X_i)$
where $\sigma(\cdot)$ is the sigmoid function. With the encoding $A\in\{0,1\}$, where 0 indicates the male baseline value, and under the approximation $\log (\pi/(1-\pi))\approx \log \pi$, 
we can write the PSE in the odds ratio scale as $\textrm{PSE}\approx\textrm{exp}(\theta^y_{a}+\theta^y_{m}\theta^m_{a}+\theta^y_{l}\theta^l_{m}\theta^m_{a}+\theta^y_{r}(\theta^r_{m}\theta^m_{a}+\theta^r_{l}\theta^l_{m}\theta^m_{a}))$.
A new instance from the test set $\{a^n,c^n,m^n,l^n,r^n\}$ is classified by using $ p(Y_{a^n}=1|c^n)=\int_{M,L,R}p(Y|a^n,c^n,M,L,R) \times p(R|a^n,c^n,M,L)p(L|a^n,c^n,M)p(M|a^n,c^n)$. The 
test accuracy obtained by constraining the PSE to lie between 0.95 and 1.05 is 72\%, compared to 82\% of the unconstrained case.

In our method, for the MMD penalization we used a two stage approach, where a factor $\beta=0$ (no penalization) was used for the first 5,000 training steps, and a factor $\beta=1000$ was used for the remaining training steps. 
For the Monte-Carlo approximation in \eqref{eq:gm}, we used $I=500$. These values were chosen based on accuracy/computational cost on the training set.

In Table \ref{tab:Adult}, we show the unfair and fair accuracy on the test set at different stages of the training, together with the corresponding MMD values for $H_m, H_l$ and $H_r$. As we can see, 
the MMD value for $H_m$ is drastically reduced from 5,000 to 8,000 and 15,000 training steps, without drastic loss in accuracy. After 20,000 training steps, the fair accuracy reduces to 
that of a dummy classifier. 
These results were obtained by performing counterfactual correction for both males and females, even if not required for males.
Fair accuracy when performing counterfactual correction only for females is instead 82.80\%, 80.05\%, 80.87\%, and 76.44\% after 5,000, 8,000, 15,000 and 20,000 training steps respectively.

\begin{figure}[t]
\begin{center}
\subfigure{
\includegraphics[height=3.7cm,width=4.1cm]{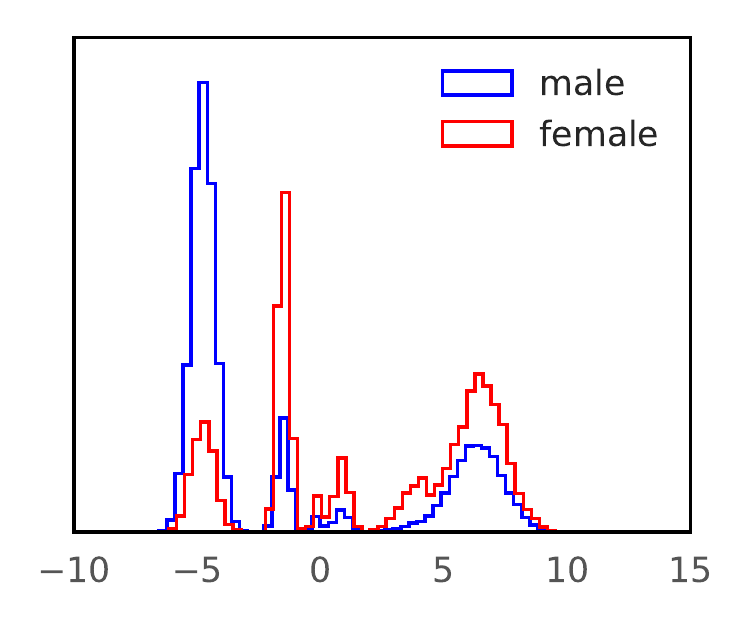}
\includegraphics[height=3.7cm,width=4.1cm]{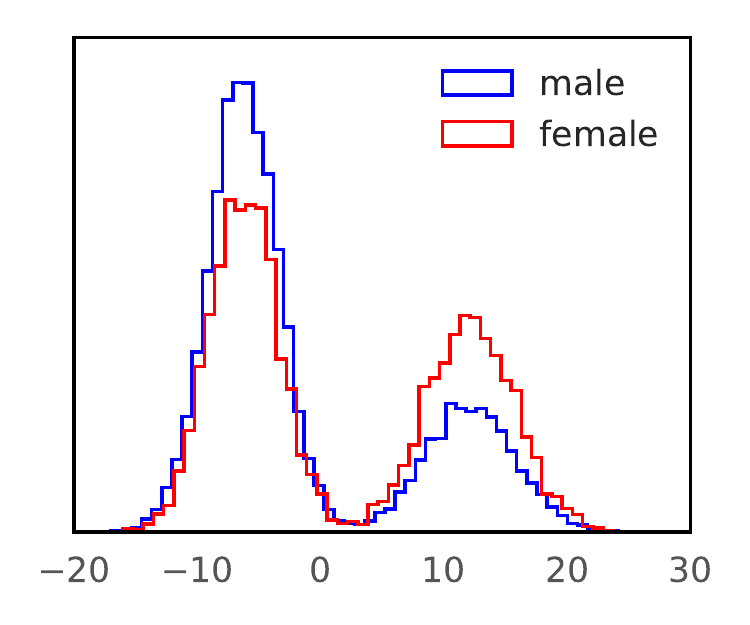}}
\subfigure{
\includegraphics[height=3.7cm,width=4.1cm]{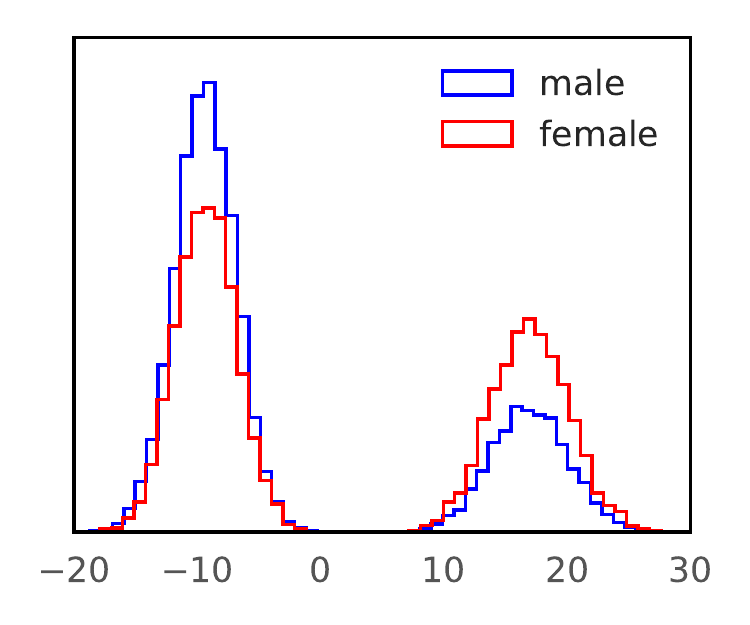}
\includegraphics[height=3.7cm,width=4.1cm]{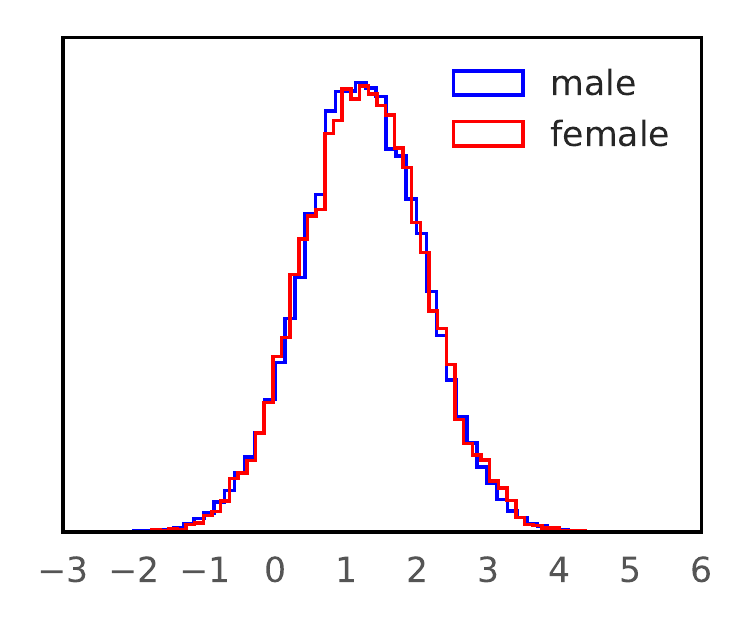}
}
\end{center}
\squeezeup
\caption{Histograms of (one dimension of) $\tilde q(H_m|A)$ after 5,000, 8,000, 15,000 and 20,000 training steps.}
\label{fig:AP}
\end{figure}

In \figref{fig:AP},
we show histograms of $\tilde q(H_m|A)$ separately for males and females for increasing numbers of training steps. The remaining variables are shown in the Appendix.
As can be seen, the addition of the MMD penalization to the variational bound for more training steps has the effect of reducing the number of modes in the posterior. 
From the evidence available, it is unclear if the shape changes are a necessary consequence of enforcing 
them to be similar, or if a simplification of the latent space is a more fundamental drawback of the MMD method. 
We leave any further investigations into such constraints for future work.

\subsection{The UCI German Credit Dataset\label{sec:GermanDataset}} 
The German Credit dataset from the UCI repository contains 20 attributes of 1,000 individuals applying for loans.
Each applicant is classified as a good or bad credit risk, \ie~as likely or not likely to repay the loan.
We assume the GCM in \figref{fig:Adult}(b), where $A$ corresponds to the protected attribute sex, $C$ to age, 
$S$ to the triple status of checking account, savings, and housing, and $R$ the duple credit amount and repayment duration. 
The attributes age, credit amount, and repayment duration are continuous, whilst checking account, savings, and housing are categorical. 
Besides the direct effect $A \rightarrow Y$, we would like to remove the effect of $A$ on $Y$ through $S$. 
We only need to introduce a hidden variable $H_s$ for $S$, as $R$ does not need to be corrected.

We divided the dataset into training and test sets of sizes 700 and 300 respectively. We used $\beta=0$ for the first 2,000 training steps, and $\beta=100$ afterward. For the Monte-Carlo approximation in \eqref{eq:gm}, we used $I=1000$.
Counterfactual correction was performed for both males and females. 

\begin{figure}[t]
\begin{center}
\subfigure{
\includegraphics[height=3.7cm,width=4.1cm]{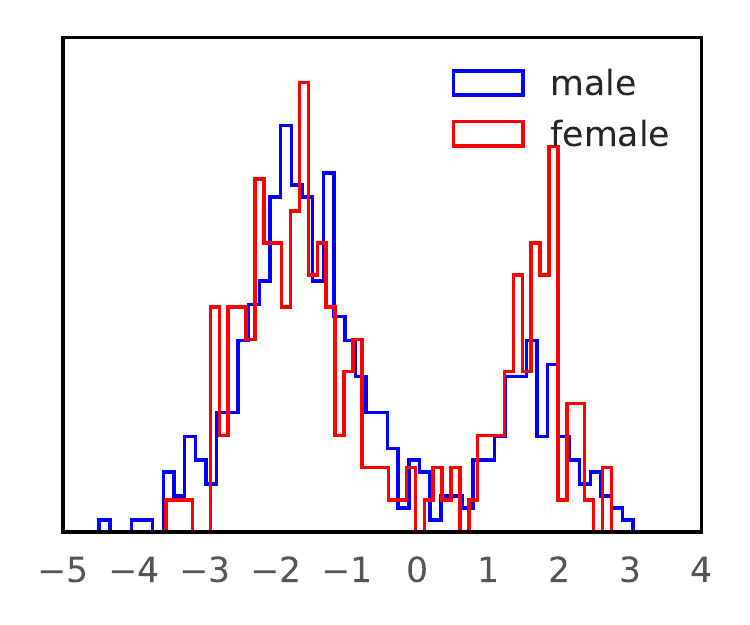}
\includegraphics[height=3.7cm,width=4.1cm]{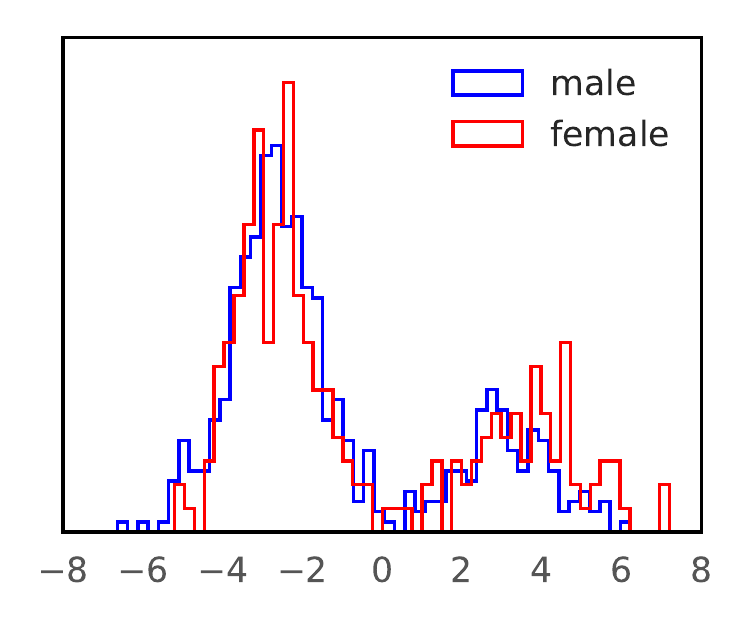}}
\end{center}
\squeezeup
\caption{Histograms of $\tilde q(H_s|A)$ after 2,000 and 8,000 training steps for one dimension of the variable housing.}
\label{fig:GP}
\end{figure}

In Table \ref{tab:German}, we show the unfair and fair test accuracy and the MMD values for $H_s$ after 2,000, 4,000, and 8,000 training steps (the results remain similar with a higher number of training steps). As we can see, unfair and fair accuracy, and 
MMD values  are similar for all iterations. This indicates that, unlike the Adult dataset, model-observations mismatch is not problematic. This is confirmed by $\tilde q(H_s|A)$; we learn a structured 
distribution which does not differ significantly for females and males. In \figref{fig:GP}, we show $\tilde q(H_s|A)$ 
for one dimension of the variable housing, which shows the most significant difference between females and males. The remaining variables are shown in the Appendix.

\section{Conclusions}

\begin{table}[]
\caption{In order columns represent: unfair test accuracy, fair test accuracy, and MMD values for $H_s$ ($\times$ 100) for the UCI German Credit dataset. Rows represent values after 2,000, 4,000, and 8,000 training steps.}
\label{tab:German}
\vskip0.25cm
\begin{center}
\begin{small}
\begin{tabular}{ccccccc}
\toprule
74.67\%  & 73.67\%  & 1.12  &  2.82  & 5.47\\
76.33\%  &  76.00\%  & 1.23  &  2.38  &  2.27\\
76.00\%  &  76.00\%  &  1.27  &  2.20  &  1.79\\
\bottomrule
\end{tabular}
\end{small}
\end{center}
\end{table}
We have introduced a latent inference-projection method to achieve path-specific counterfactual fairness which simplifies, generalizes and outperforms previous literature.
A fair decision is achieved by correcting the variables that are descendants of the protected attribute along unfair pathways, rather than by imposing constraints on the model parameters.
This enables us to retain fair information contained in the problematic descendants and to leave unaltered the underlying data-generation mechanism. In the future, we plan to investigate 
alternative techniques to MMD for enforcing independence between the latent space and the sensitive attribute.

\bibliography{PSCF}
\bibliographystyle{icml2018}

\appendix
\onecolumn
\section{Identifiability of PSE}
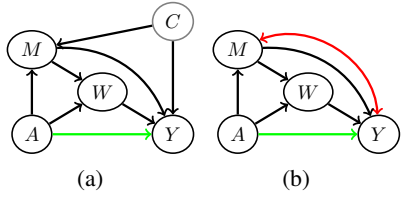
\begin{figure}[t]
\begin{center}
\subfigure[]{
\scalebox{0.75}{
\begin{tikzpicture}[dgraph]
\node[hcont] (C) at (6,1.5) {$C$};
\node[ocont] (M) at (3.5,1) {$M$};
\node[ocont] (W) at (4.75,0.25) {$W$};
\node[ocont] (X) at (3.5,-0.5) {$A$};
\node[ocont] (Y) at (6,-0.5) {$Y$};
\draw[line width=1.15pt](C)--(M);\draw[line width=1.15pt](C)--(Y);
\draw[line width=1.15pt](X)--(M);\draw[line width=1.15pt](X)--(W);\draw[line width=1.15pt,green](X)--(Y);\draw[line width=1.15pt](M)--(W);\draw[line width=1.15pt](W)--(Y);
\draw[line width=1.15pt](M)to [bend right=-35](Y);
\end{tikzpicture}}}
\subfigure[]{
\scalebox{0.75}{
\begin{tikzpicture}[dgraph]
\node[ocont] (M) at (3.5,1) {$M$};
\node[ocont] (W) at (4.75,0.25) {$W$};
\node[ocont] (X) at (3.5,-0.5) {$A$};
\node[ocont] (Y) at (6,-0.5) {$Y$};
\draw[<->,line width=1.15pt,red](M)to [bend right=-55](Y);
\draw[line width=1.15pt](X)--(M);\draw[line width=1.15pt](X)--(W);\draw[line width=1.15pt,green](X)--(Y);\draw[line width=1.15pt](M)--(W);\draw[line width=1.15pt](W)--(Y);
\draw[line width=1.15pt](M)to [bend right=-35](Y);
\end{tikzpicture}}}
\caption{(a): GCM with an unobserved confounder $C$ indicated with a gray node. (b): ADMG corresponding to (a). The causal effect along the green path $A \rightarrow Y$ cannot be identified by only using observed variables.}
\label{fig:PSEAppendix}
\end{center}
\end{figure}
We summarize the method described in \citet{shpitser13counterfactual} to graphically establish whether a PSE is identifiable.

\paragraph{Acyclic Directed Mixed Graph (ADMG):} An ADMG is a causal graph containing two kinds of links, directed links (either green or black depending on whether we are interested in the corresponding causal path), and red bidirected links, 
indicating the presence of an unobserved common cause. The ADMG corresponding to \figref{fig:PSEAppendix}(a) is given by \figref{fig:PSEAppendix}(b).

\paragraph{District:} The set of nodes in an ADMG that are reachable from $A$ through bidirected paths is called the district of $A$. For example, the district of $Y$ in \figref{fig:PSEAppendix}(b) is $\{M,Y\}$.

\paragraph{Recanting District:} Let ${\cal G}$ be an ADMG, and $\pi$ a subset of causal paths which start in $A$ and end in $Y$. 
Let $\cal{V}$ be the set of potential causes of $Y$ that differ from $A$ and that influence $Y$ through causal paths that do not intersect $A$. 
Let ${\cal G}_{{\cal V}}$ be the subgraph of ${\cal G}$ containing only the nodes in $\cal{V}$.
A district $D$ in ${\cal G}_{{\cal V}}$ is called the recanting district for the effect of $A$ on $Y$ along the paths in $\pi$ if there exist nodes $X_i,X_j\in D$ such that there is a causal path $A\rightarrow X_i \rightarrow \ldots \rightarrow Y$ $\in\pi$
and a causal path $A\rightarrow X_j \rightarrow \ldots \rightarrow Y$ $\notin\pi$. If ${\cal G}_{{\cal V}}$ contains a recanting district for $\pi$, then the effect along $\pi$ is non-identifiable.

For example, the set ${\cal V}$ in \figref{fig:PSEAppendix}(b) is $\{M,W,Y\}$. The districts in ${\cal G}_{{\cal V}}$ are $\{M,Y\}$. This district is recanting for the effect along $A\rightarrow Y$, as $A \rightarrow Y \in \pi$, whilst $A \rightarrow M \rightarrow Y \notin \pi$.
(This district is not recanting for the effect along $A\rightarrow W \rightarrow Y$.)

\section{Latent-Variable Conditional Distribution}
Consider the GCM in Fig. 3(a), corresponding to Eq. (3) with the addition of a Gaussian latent variable $H_m\sim {\cal N}(\theta^h,\sigma^2_h)$ in the equation for $M$.
The joint distribution $p(Z=\{Y,L,M,C,H_m\}|A)$ is Gaussian with exponent proportional to $-\frac{1}{2}(Z'NZ -2n)$
with 
\begin{align*}
&N=
\left(
\begin{array}{ccccc}
\frac{1}{\sigma^2_y} & -\frac{\theta^y_l}{\sigma^2_y} & -\frac{\theta^y_m}{\sigma^2_y}  & -\frac{\theta^y_c}{\sigma^2_y}  & 0\\
-\frac{\theta^y_l}{\sigma^2_y} & \frac{1}{\sigma^2_l}+\frac{(\theta^y_l)^2}{\sigma^2_y} & \frac{\theta^y_l\theta^y_m}{\sigma^2_y} - \frac{\theta^l_m}{\sigma^2_l} & \frac{\theta^y_l\theta^y_c}{\sigma^2_y}  - \frac{\theta^l_c}{\sigma^2_l} & 0\\\
-\frac{\theta^y_m}{\sigma^2_y} & \frac{\theta^y_l\theta^y_m}{\sigma^2_y}  - \frac{\theta^l_m}{\sigma^2_l} & \frac{1}{\sigma^2_m}+\frac{(\theta^y_m)^2}{\sigma^2_y}+\frac{(\theta^l_m)^2}{\sigma^2_l} & \frac{\theta^y_m\theta^y_c}{\sigma^2_y} + \frac{\theta^l_m\theta^l_c}{\sigma^2_l} - \frac{\theta^m_c}{\sigma^2_m} & -\frac{\theta^m_h}{\sigma^2_{m}}\\
-\frac{\theta^y_c}{\sigma^2_y} &  \frac{\theta^y_l\theta^y_c}{\sigma^2_y} - \frac{\theta^l_c}{\sigma^2_l}  & \frac{\theta^y_m\theta^y_c}{\sigma^2_y} + \frac{\theta^l_m\theta^l_c}{\sigma^2_l} - \frac{\theta^m_c}{\sigma^2_m}  &\frac{1}{\sigma^2_c}+\frac{(\theta^y_c)^2}{\sigma^2_y}+\frac{(\theta^l_c)^2}{\sigma^2_l}+\frac{(\theta^m_c)^2}{\sigma^2_m} &  \frac{\theta^m_c\theta^m_h}{\sigma^2_{m}}  \\
0 & 0 & -\frac{\theta^m_h}{\sigma^2_{m}} &  \frac{\theta^m_c\theta^m_h}{\sigma^2_{m}} & \frac{1}{\sigma^2_{h}} + \frac{(\theta^m_h)^2}{\sigma^2_m} \\
\end{array} \right)\,,
\\[5pt]
&n=\left(
\begin{array}{c}
\frac{\theta^y+\theta^y_aA}{\sigma^2_y}\\
-\frac{\theta^y_l(\theta^y+\theta^y_aA)}{\sigma^2_y} +\frac{\theta^l+\theta^l_aA}{\sigma^2_l}\\
-\frac{\theta^y_m(\theta^y+\theta^y_aA)}{\sigma^2_y} -\frac{\theta^l_m(\theta^l+\theta^l_aA)}{\sigma^2_l}+\frac{\theta^m+\theta^m_aA}{\sigma^2_m}\\
-\frac{\theta^y_c(\theta^y+\theta^y_aA)}{\sigma^2_y} -\frac{\theta^l_c(\theta^l+\theta^l_aA)}{\sigma^2_l}-\frac{\theta^m_c(\theta^m+\theta^m_aA)}{\sigma^2_m}+\frac{\theta^c}{\sigma^2_c}\\
-\frac{\theta^m_h(\theta^m+\theta^m_aA)}{\sigma^2_m}+\frac{\theta^h}{\sigma^2_h}\\
\end{array} \right)\,.
\end{align*}
The Gaussian conditional $p(H_m| A, C, M, L)$ can be computed through the formulas for Gaussian marginalization and conditioning.
\begin{figure}[t]
\begin{center}
\subfigure{
\includegraphics[height=3cm,width=3.4cm]{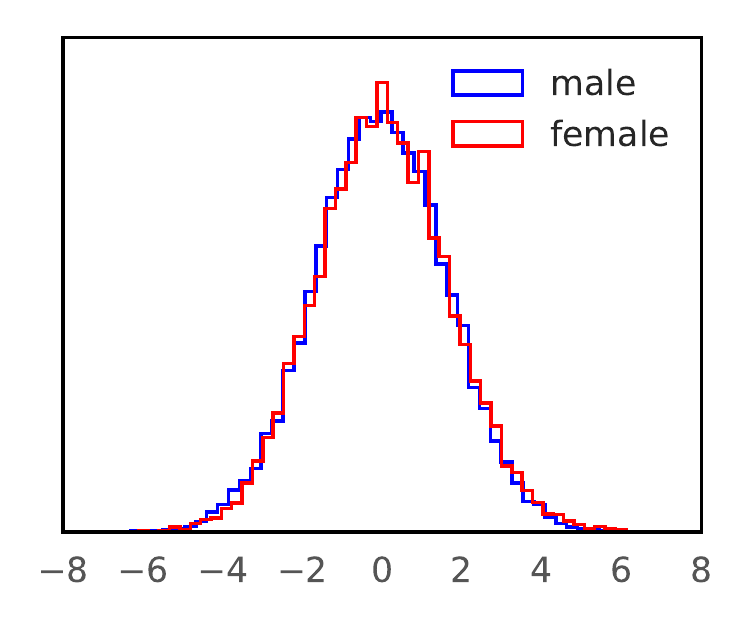}
\includegraphics[height=3cm,width=3.4cm]{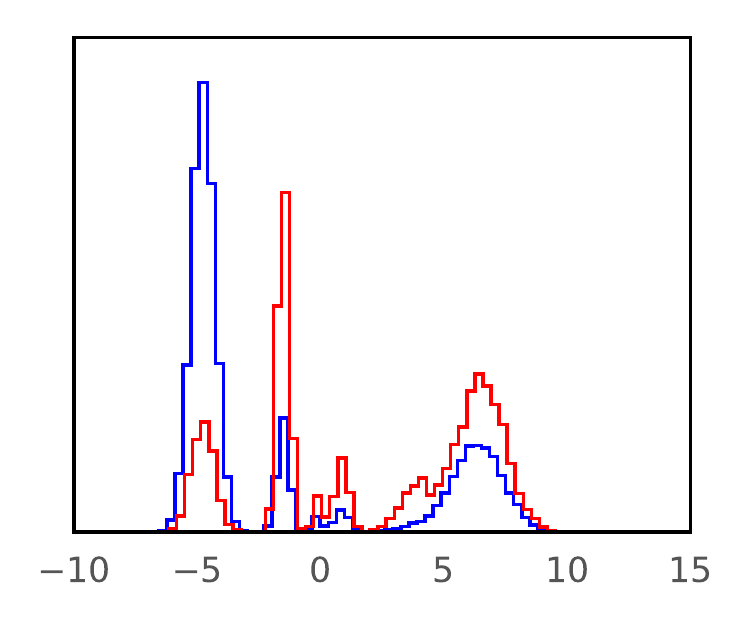}
\includegraphics[height=3cm,width=3.4cm]{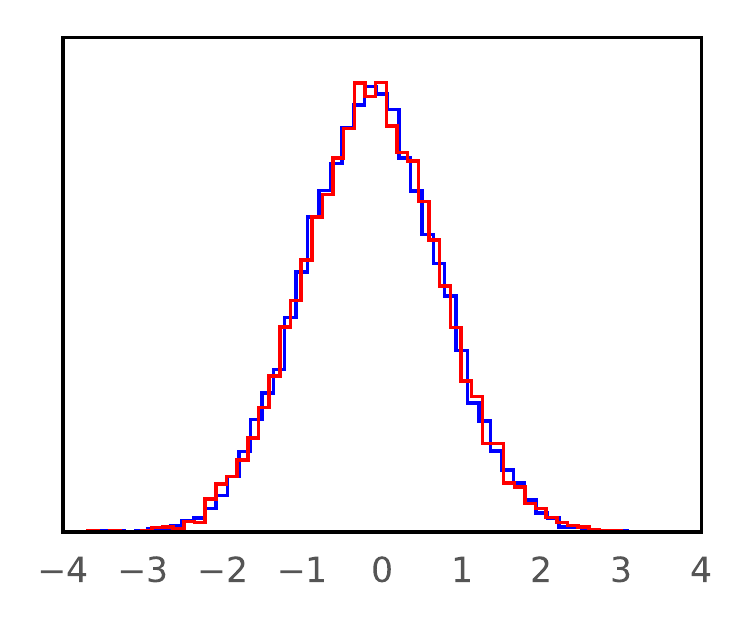}
\includegraphics[height=3cm,width=3.4cm]{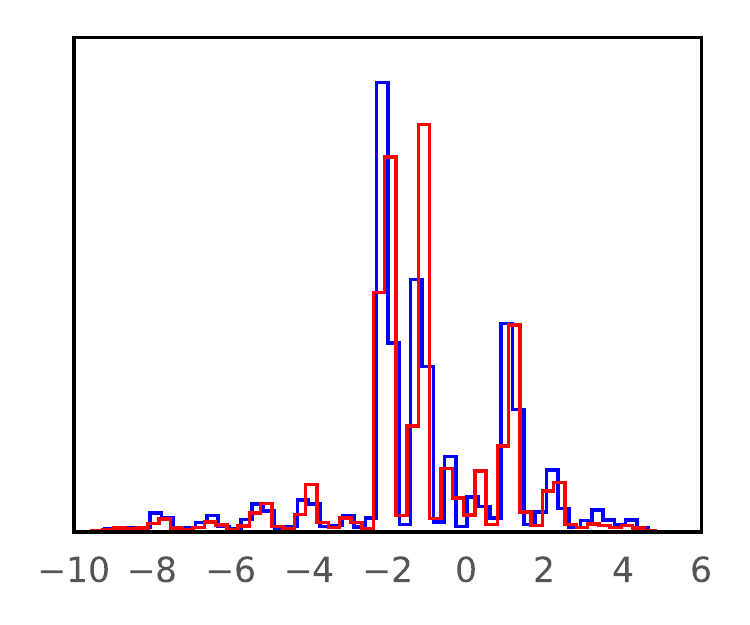}
\includegraphics[height=3cm,width=3.4cm]{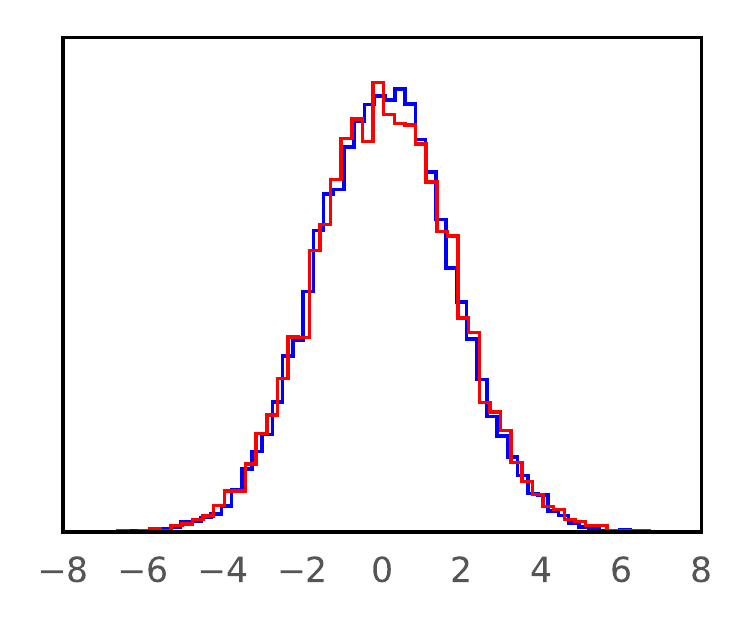}
}
\renewcommand{\thesubfigure}{(a)}
\subfigure[]{
\includegraphics[height=3cm,width=3.4cm]{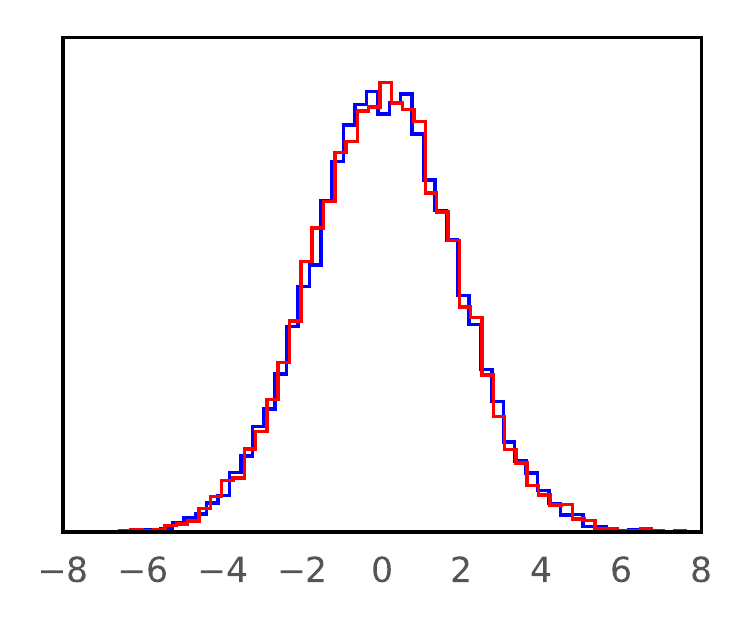}
\includegraphics[height=3cm,width=3.4cm]{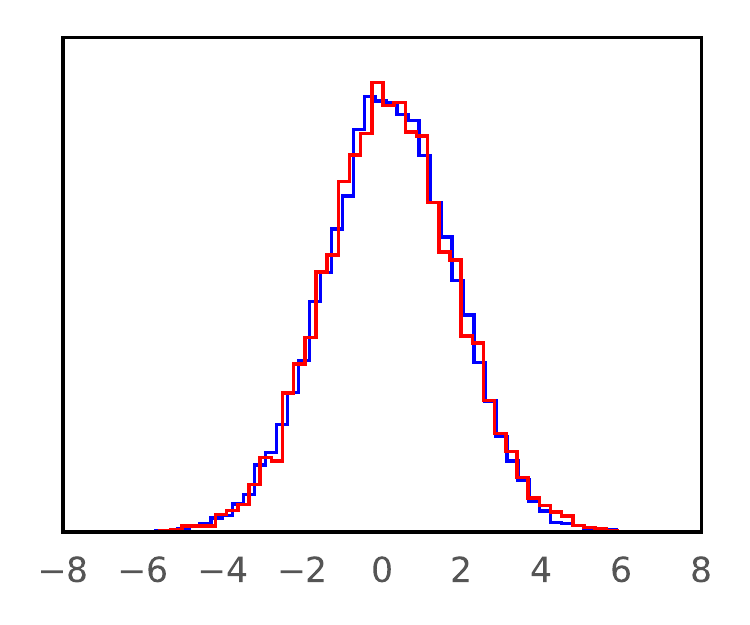}
\includegraphics[height=3cm,width=3.4cm]{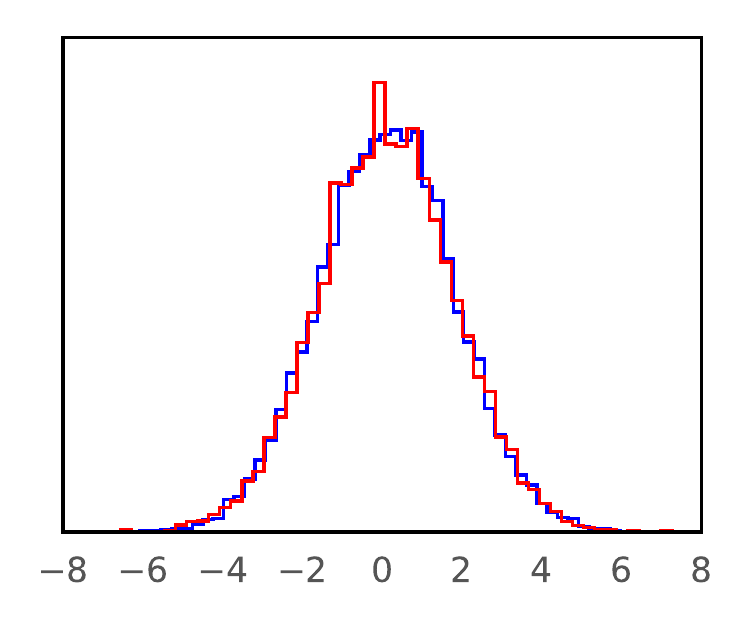}
\includegraphics[height=3cm,width=3.4cm]{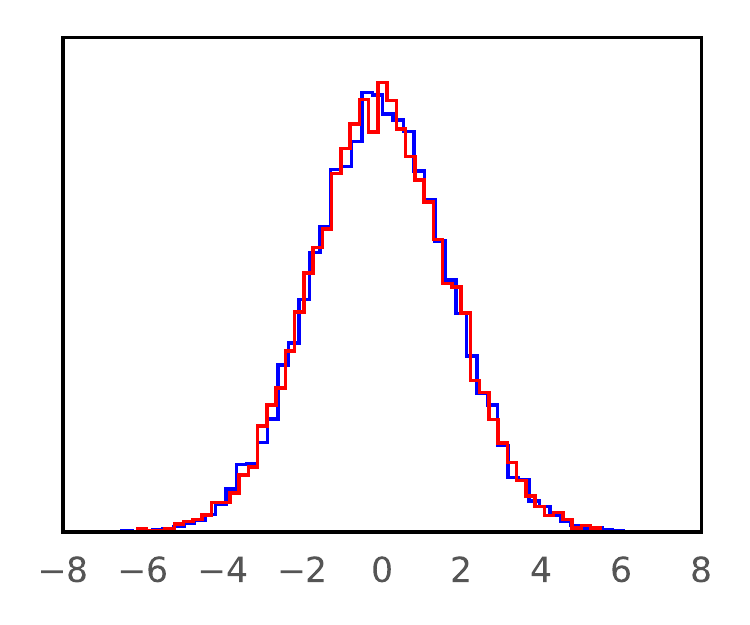}
\includegraphics[height=3cm,width=3.4cm]{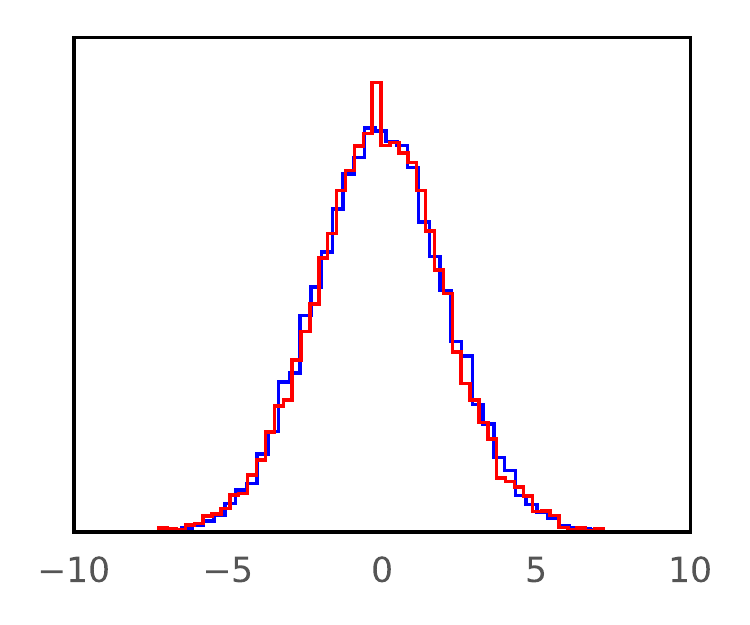}
}
\end{center}
\begin{center}
\subfigure{
\includegraphics[height=3cm,width=3.4cm]{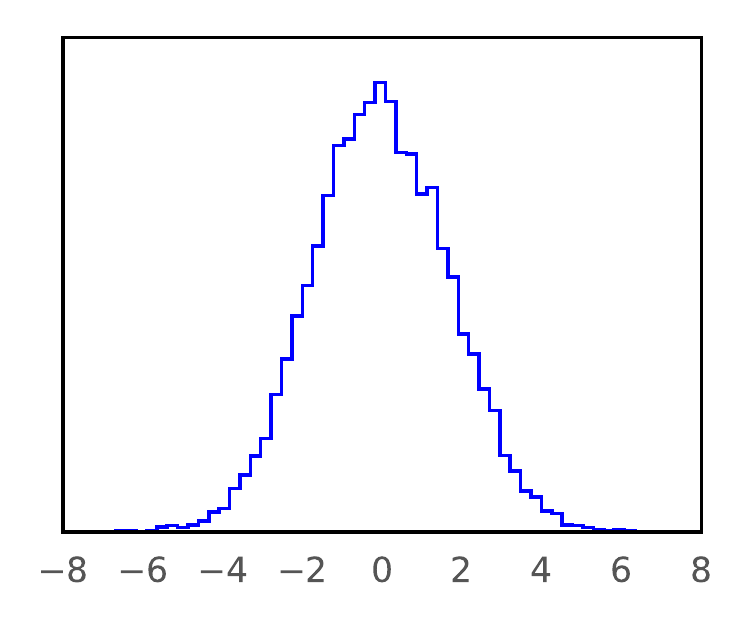}
\includegraphics[height=3cm,width=3.4cm]{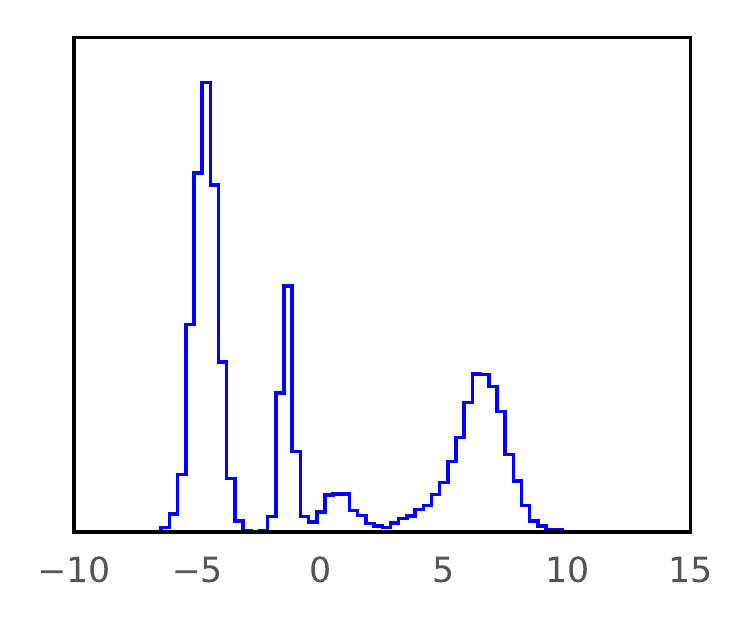}
\includegraphics[height=3cm,width=3.4cm]{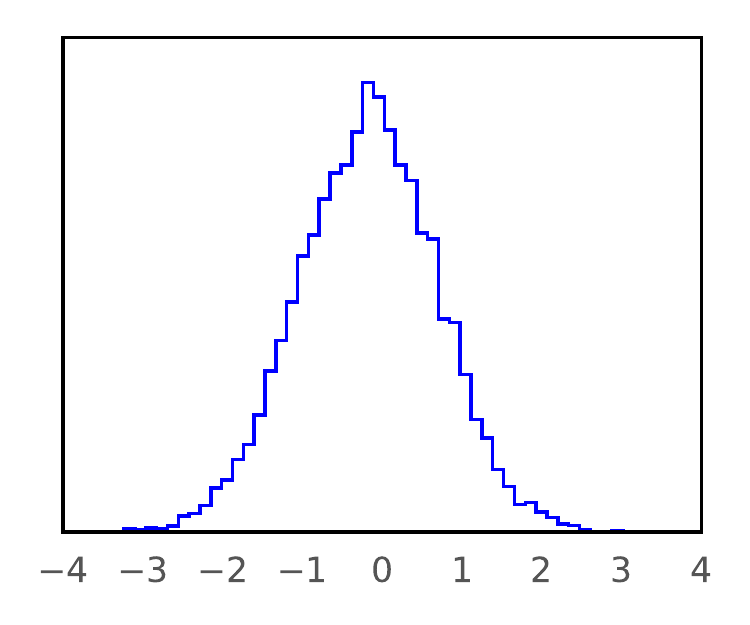}
\includegraphics[height=3cm,width=3.4cm]{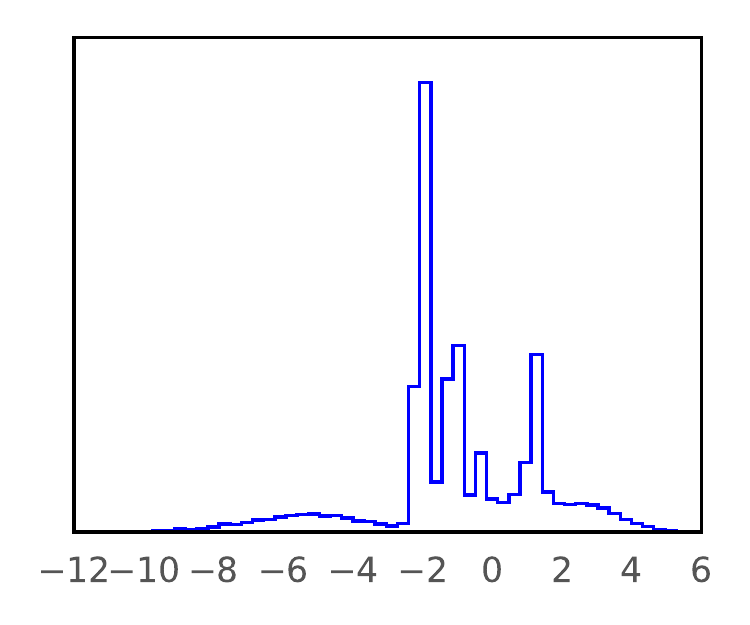}
\includegraphics[height=3cm,width=3.4cm]{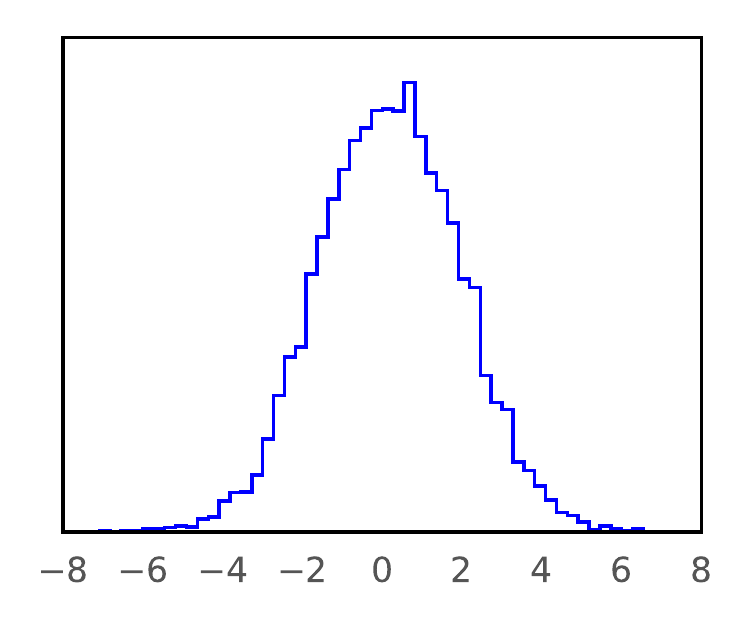}
}
\renewcommand{\thesubfigure}{(b)}
\subfigure[]{
\includegraphics[height=3cm,width=3.4cm]{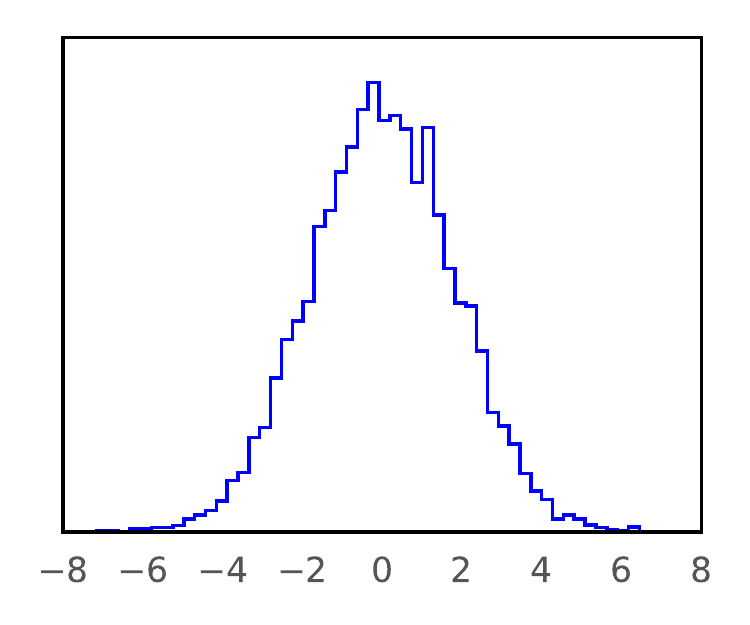}
\includegraphics[height=3cm,width=3.4cm]{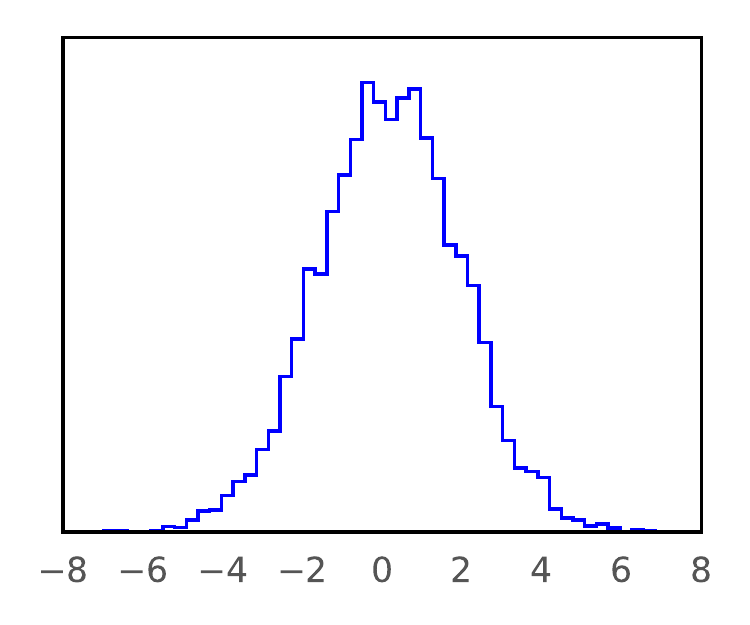}
\includegraphics[height=3cm,width=3.4cm]{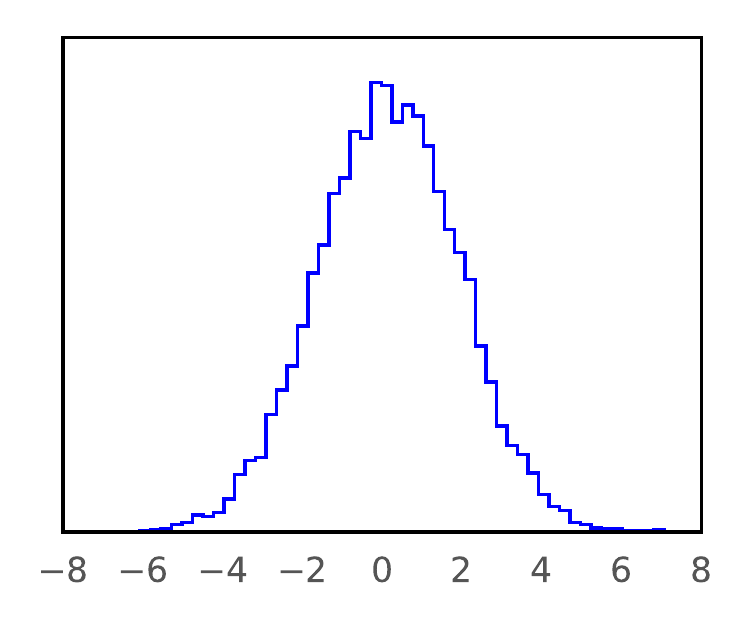}
\includegraphics[height=3cm,width=3.4cm]{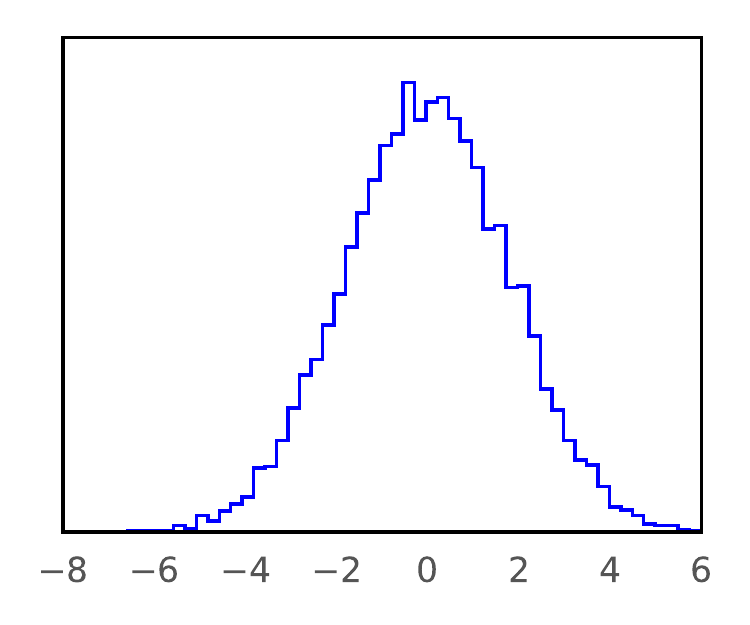}
\includegraphics[height=3cm,width=3.4cm]{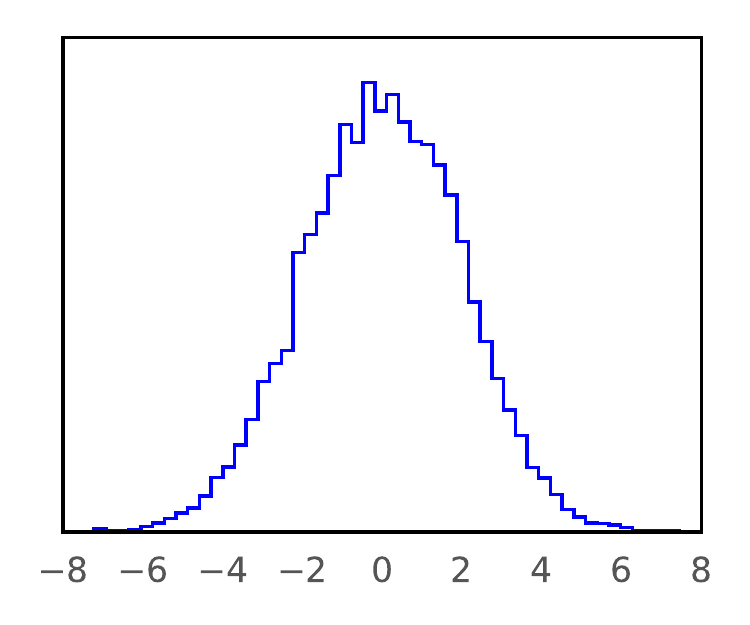}
}
\end{center}
\caption{(a): Histograms of $\tilde q(H_m|A)$ (two-dimensional), $\tilde q(H_l|A)$ (two-dimensional), and $\tilde q(H_r|A)$ (six-dimensional) after 5,000 training steps. (b): Prior distributions $p(H_m)$, $p(H_l)$, and $p(H_r)$ corresponding to mixtures of ten two-dimensional Gaussians.}
\label{fig:APAppendix}
\end{figure}

\section{Experimental Details}
For all datasets, as the prior distribution $p$ for each latent variable we used a mixture of two-dimensional Gaussians with ten mixture components and diagonal covariances.
As the variational posterior distribution $q$ we used a two-dimensional Gaussian with diagonal covariance, with 
means and log variances obtained as the outputs of a neural network with two linear layers of size 20 and tanh activation, followed by a linear layer. 
In the conditional distributions, $f_{\theta}$ was a neural network with one linear layer of size 100 with tanh activation, followed by a linear layer. 
The outputs were Gaussian means for continuous variables and logits for categorical variables.
We used the Adam optimizer \cite{kingma15adam} with learning rate 0.01, mini-batch size 128, and default values $\beta_1 = 0.9$, $\beta_2 = 0.999$, and $\epsilon=1e$-8.

\subsection{UCI Adult Dataset}
In \figref{fig:APAppendix} we show histograms for prior and posterior distributions in the latent space. 

\subsection{UCI German Credit Dataset}

\begin{figure}[t]
\begin{center}
\subfigure{
\includegraphics[height=3cm,width=3.4cm]{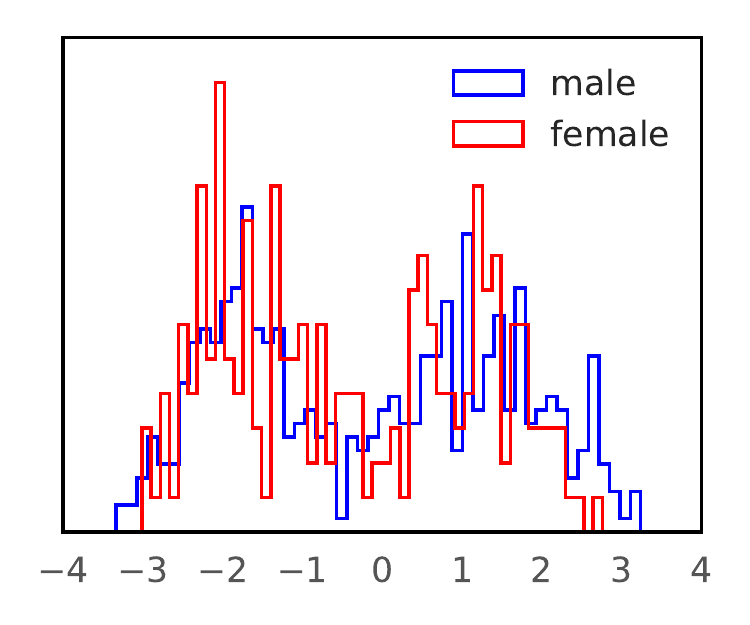}
\includegraphics[height=3cm,width=3.4cm]{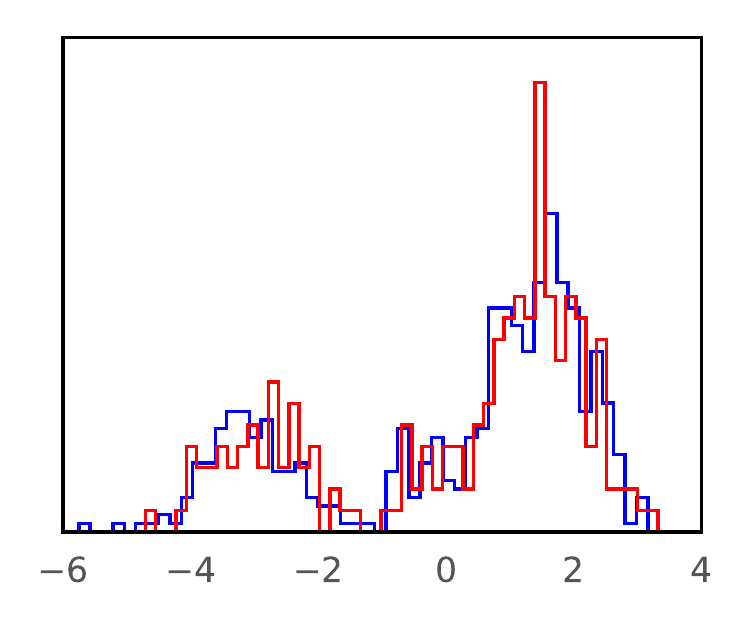}
\includegraphics[height=3cm,width=3.4cm]{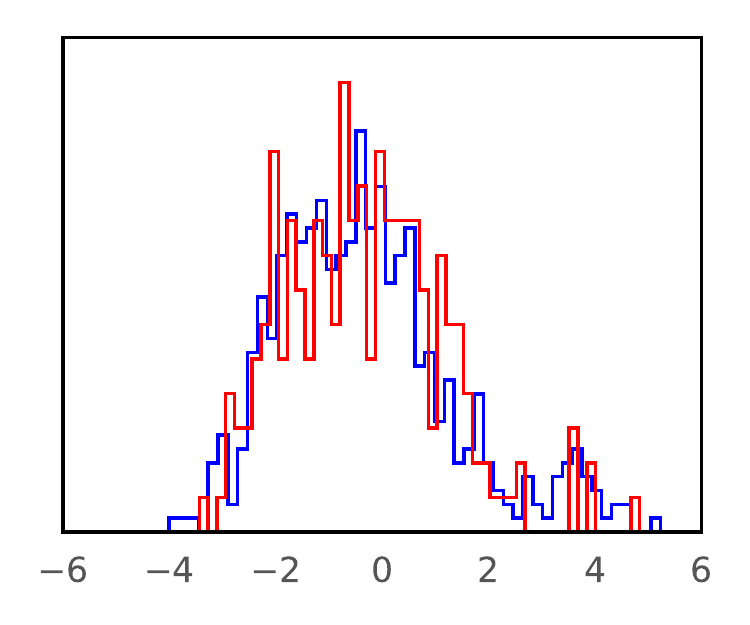}
\includegraphics[height=3cm,width=3.4cm]{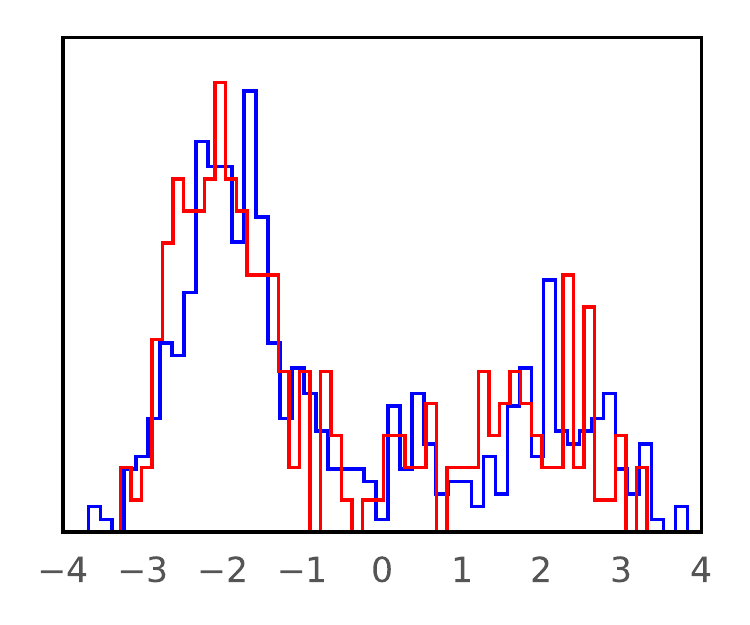}
\includegraphics[height=3cm,width=3.4cm]{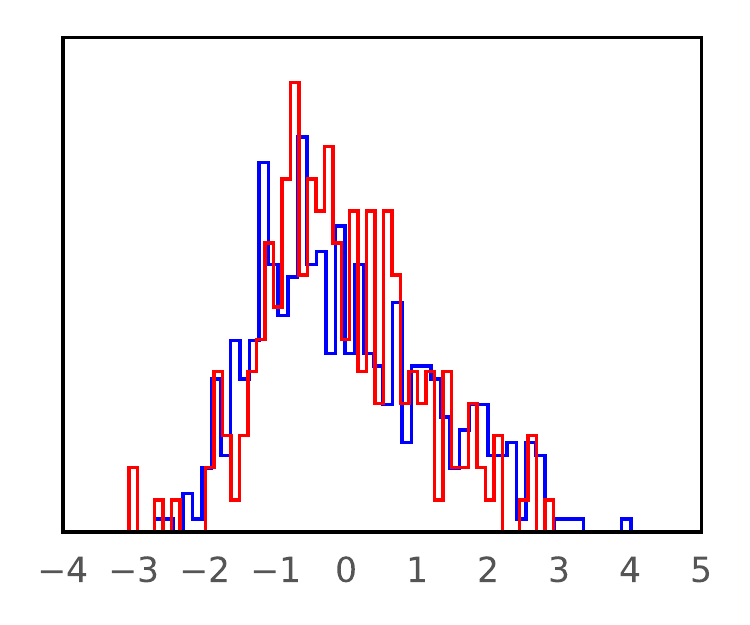}
}
\subfigure{
\includegraphics[height=3cm,width=3.4cm]{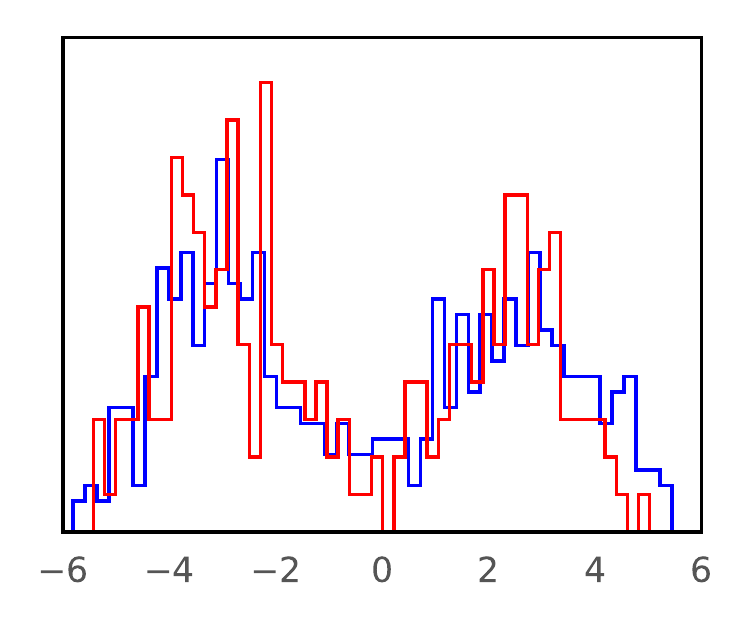}
\includegraphics[height=3cm,width=3.4cm]{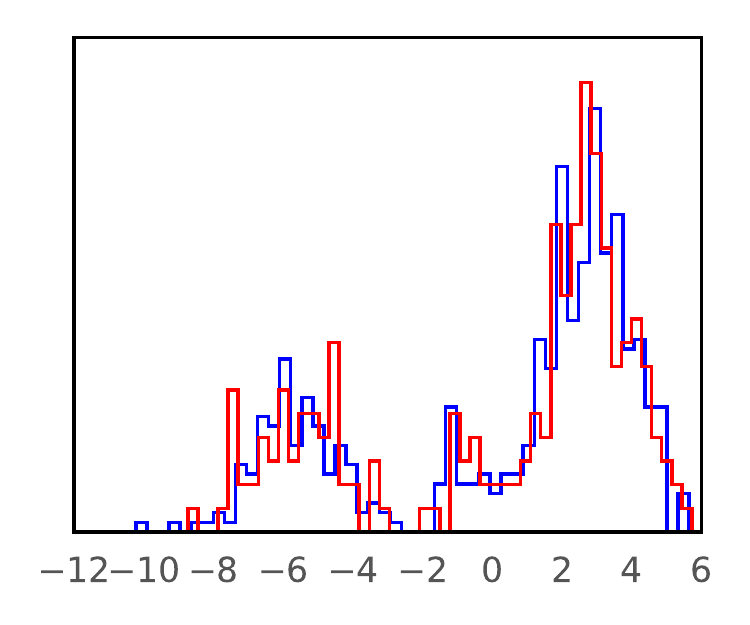}
\includegraphics[height=3cm,width=3.4cm]{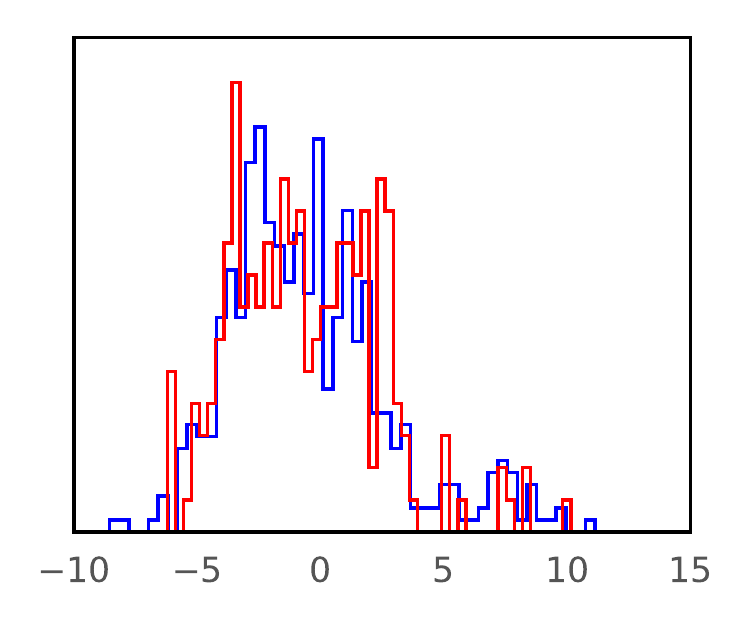}
\includegraphics[height=3cm,width=3.4cm]{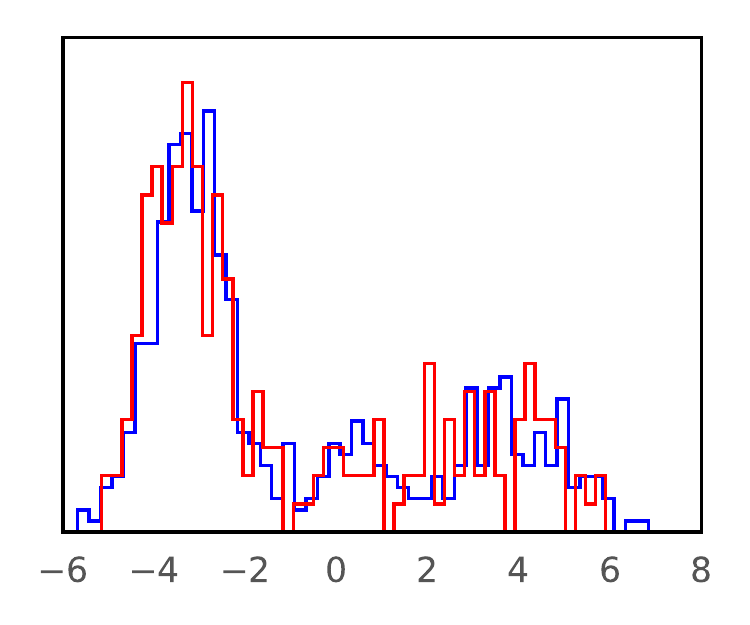}
\includegraphics[height=3cm,width=3.4cm]{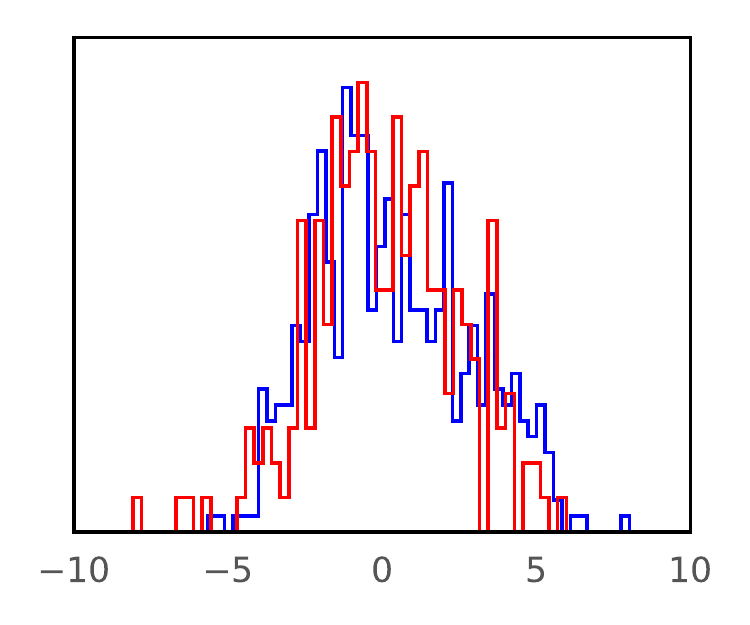}
}
\end{center}
\caption{Histograms of $\tilde q(H_s|A)$ after 2,000 (first row) and 8,000 (second row) training steps. From left to right: status of checking account (two dimensions), savings (two dimensions), and housing (one dimension).}
\label{fig:GPAppendix}
\end{figure}
In \figref{fig:GPAppendix} we show histograms for posterior distributions in the latent space.

\end{document}